\documentclass{zlab}
\usepackage[iclrfont]{gaze_arxiv}
\titlespacing*{\subsubsection}{0pc}{1ex plus1pt minus0.5pt}{0pt}

\usepackage{amsmath,amsfonts,bm}

\def\eqref#1{equation~\ref{#1}}

\def\1{\bm{1}}

\DeclareMathAlphabet{\mathsfit}{\encodingdefault}{\sfdefault}{m}{sl}
\SetMathAlphabet{\mathsfit}{bold}{\encodingdefault}{\sfdefault}{bx}{n}

\usepackage{hyperref}
\usepackage{etoc}
\usepackage{url}
\usepackage{xcolor}
\usepackage{colortbl}
\usepackage{graphicx}
\usepackage{capt-of}
\usepackage{booktabs}
\usepackage{float}
\usepackage{placeins}       
\usepackage{wrapfig}
\usepackage{needspace}
\usepackage{tcolorbox}
\tcbuselibrary{breakable}
\tcbuselibrary{skins}
\definecolor{findingborder}{HTML}{005B5B}
\definecolor{findingbackground}{HTML}{F2F2F2}
\newtcolorbox[auto counter]{findingbox}{%
    enhanced,
    colback=findingbackground,
    colframe=findingborder,
    boxrule=0.65pt,
    arc=5pt,
    boxsep=0pt,
    left=11pt, right=11pt, top=7pt, bottom=7pt,
    before skip=8pt, after skip=8pt,
    shadow={2pt}{-2pt}{0pt}{black!12},
    fontupper=\normalsize,
    before upper={\textsf{\textbf{Finding~\thetcbcounter:}}\enspace}
}

\definecolor{promptbg}{HTML}{F9F9F9}
\definecolor{promptcodebg}{HTML}{EFEFEF}
\tcbset{promptstylefmt/.style={
    breakable, colback=promptbg, boxrule=0.9pt,
    left=10pt, right=10pt, top=8pt, bottom=8pt,
    fontupper=\small\sffamily,
    fonttitle=\sffamily\fontseries{bx}\selectfont,
    toptitle=5pt, bottomtitle=5pt, lefttitle=10pt,
    boxsep=0pt, sharp corners
}}
\newtcolorbox[auto counter,number within=section]{promptbox}[2]{
    promptstylefmt,title={Listing \thetcbcounter: #1},label={#2}
}
\newcommand{\pbold}[1]{{\sffamily\bfseries #1}}
\newcommand{\pcode}[1]{{\small\ttfamily\colorbox{promptcodebg}{\strut #1}}}
\newcommand{\pboxed}[1]{\pcode{\textbackslash boxed\{#1\}}}
\newcommand{\pvar}[1]{{\small\ttfamily\{#1\}}}
\newcommand{\promptrule}{\par\vspace{4pt}\noindent\rule{\linewidth}{0.4pt}\par\vspace{4pt}}
\newenvironment{promptitemize}{%
    \begin{list}{\textbullet}{%
        \setlength{\topsep}{2pt}\setlength{\itemsep}{1pt}%
        \setlength{\parsep}{0pt}\setlength{\leftmargin}{1.5em}%
    }%
}{\end{list}}
\newcounter{promptenumi}
\newenvironment{promptenumerate}{%
    \begin{list}{\arabic{promptenumi}.}{%
        \usecounter{promptenumi}%
        \setlength{\topsep}{2pt}\setlength{\itemsep}{1pt}%
        \setlength{\parsep}{0pt}\setlength{\leftmargin}{2em}%
    }%
}{\end{list}}

\title{%
    Learning Grounded Reasoning from Human Gaze\par
    \vspace{-0.5cm}
}

\author{%
    \vspace{.2cm}
    \parbox{\textwidth}{\centering
        Gabriel Sarch$^{1,*}$ \quad
        Declan Campbell$^{1,2,*}$ \quad
        Nicholas Budny$^{1,*}$ \quad
        Kia Ghods$^{1}$ \\
        \vspace{-0.0em}\Authfont
        Katerina Fragkiadaki$^{3}$ \quad
        Jonathan D. Cohen$^{1,\dagger}$ \quad
        Brenden M. Lake$^{1,\dagger}$ \quad
        Taylor W. Webb$^{1,\dagger}$
    }
    \\
    \vspace{.0cm}
    {\normalfont\fontsize{11}{15}\selectfont
        $^{1}$Princeton University \quad $^{2}$Perceptron AI \quad
        $^{3}$Carnegie Mellon University}
    \vspace{.5cm}
}
\correspondingauthor{\shortstack[l]{\, $^{*}$ Equal contribution\hspace{1em} $^{\dagger}$ Joint supervision}}

\newcommand{\codeurl}{https://github.com/Gabesarch/GGR}
\newcommand{\dataurl}{https://huggingface.co/datasets/gsarch/GGR-raw-human-gaze-dataset}
\newcommand{\modelsurl}{https://huggingface.co/collections/gsarch/ggr}
\newcommand{\resourceicon}[1]{%
    \raisebox{-1.3pt}{#1}%
}
\newcommand{\resourcelink}[2]{%
    \ifx#1\empty
        \textcolor{resourcelinkcolor}{\textbf{#2}}%
    \else
        \href{#1}{\textbf{#2}}%
    \fi
}
\newcommand{\resourcerow}{%
    \begin{center}
    \small
    \resourceicon{\includegraphics[height=1.05em]{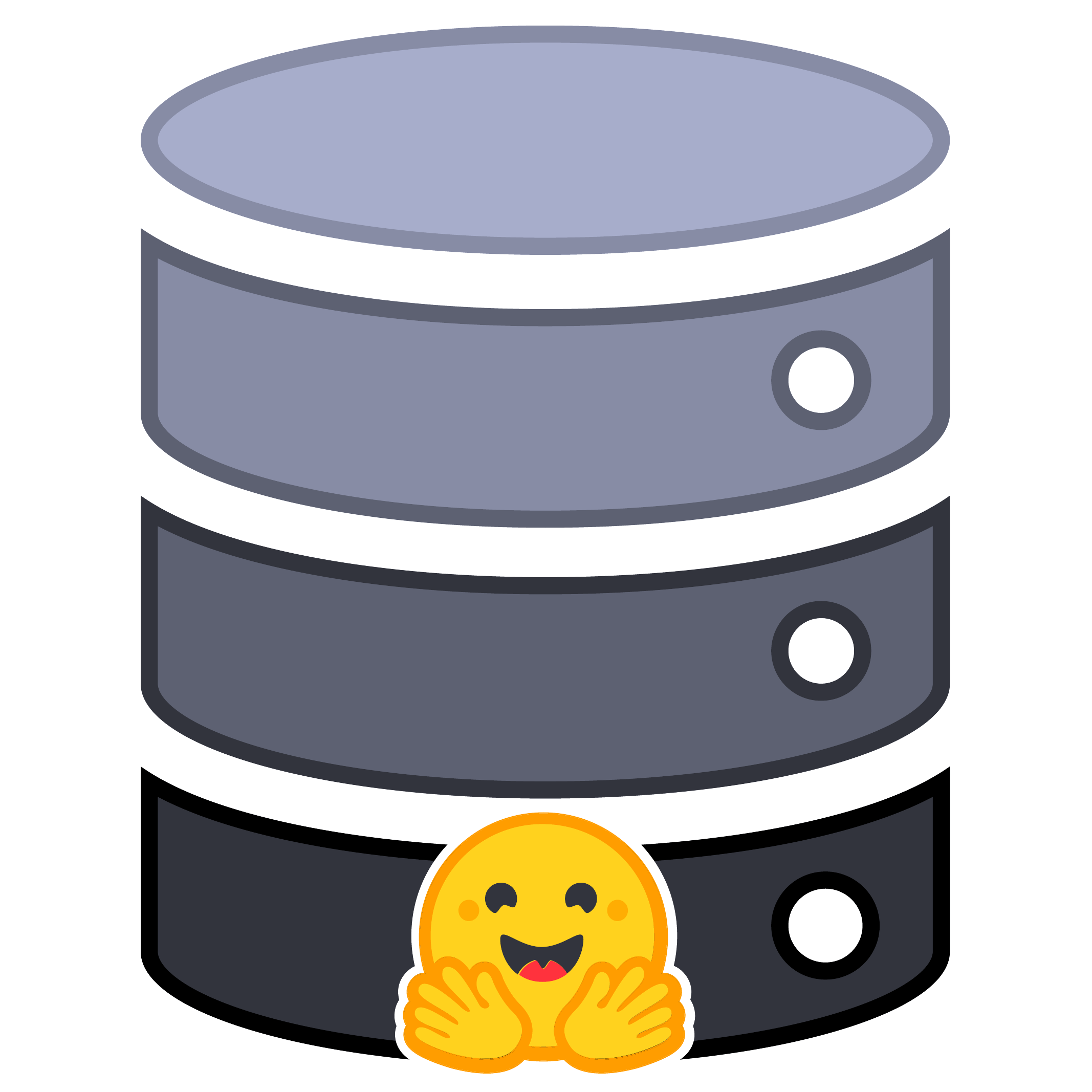}}\,\resourcelink{\dataurl}{Data}
    \qquad
    \resourceicon{\includegraphics[height=1.05em]{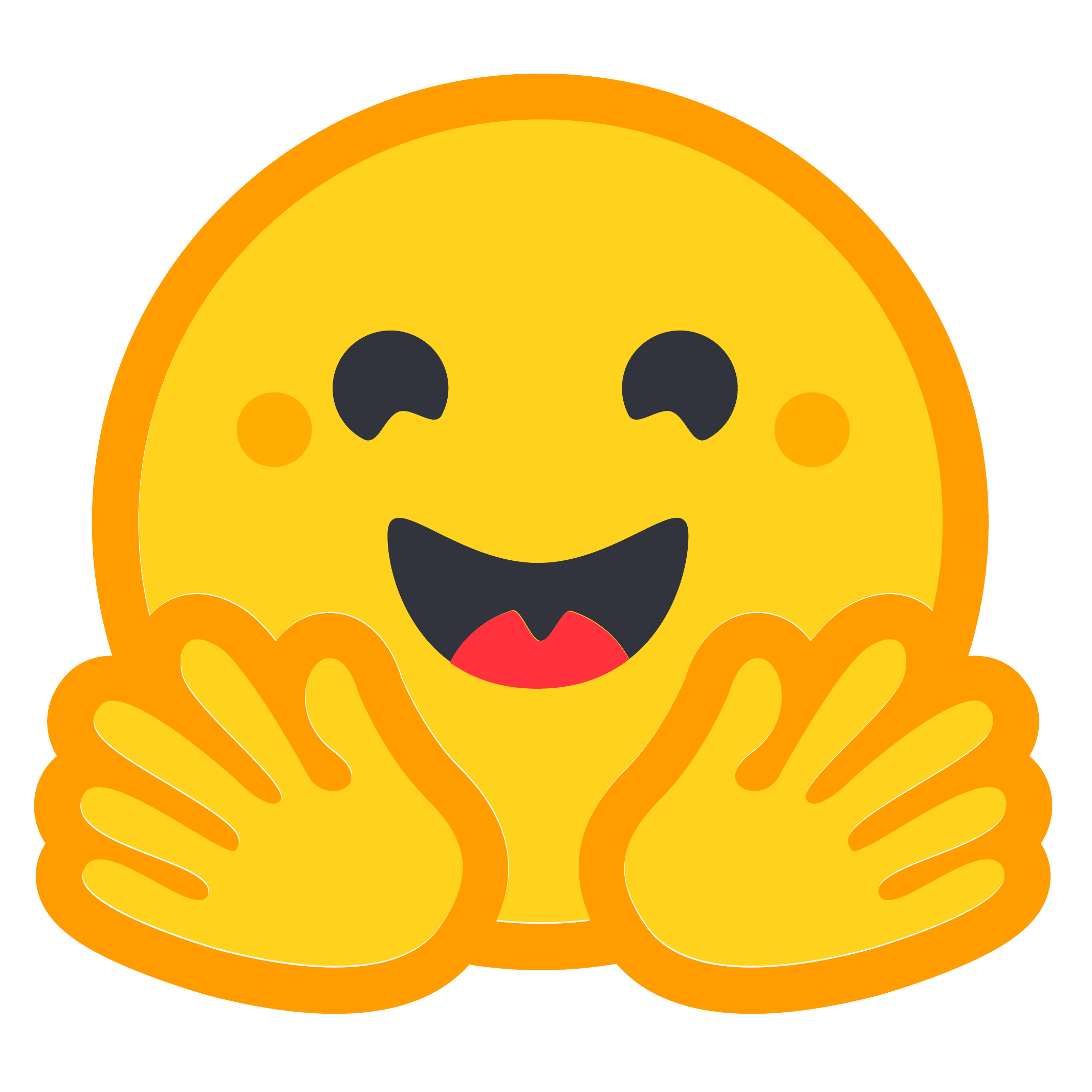}}\,\resourcelink{\modelsurl}{Models}
    \qquad
    \resourceicon{\includegraphics[height=1.05em]{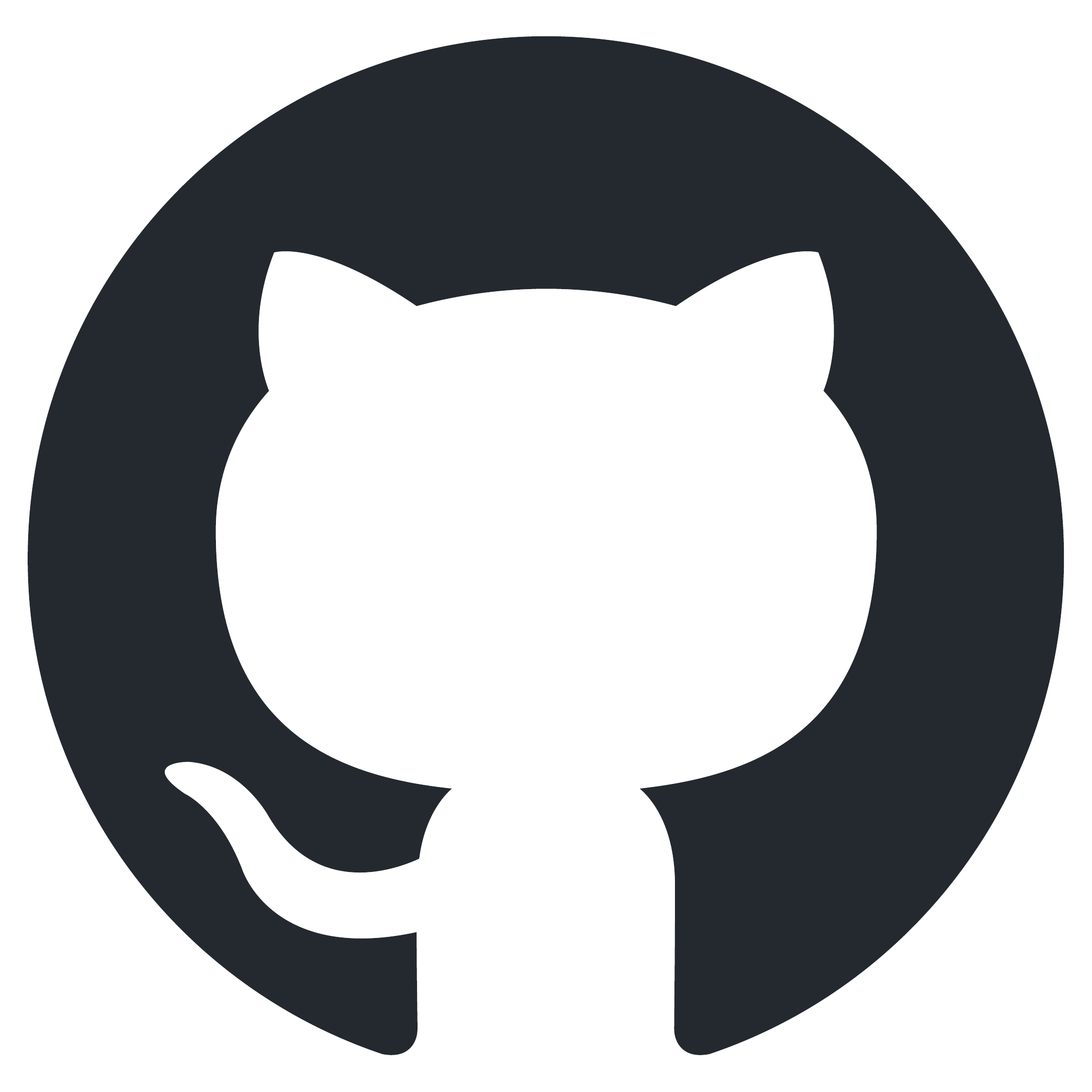}}\,\resourcelink{\codeurl}{Code}
    \end{center}%
}

\newcommand{\model}[1]{GGR}
\newcommand{\modelsft}{\model{}-SFT}
\newcommand{\modelrl}{\model{}-RL}

\newcommand{\includevisualgroundingaudit}{}

\begin{document}

\begingroup
\let\raggedright\centering
\maketitle
\endgroup

\newcommand{\teasergraphicwidth}{0.98\linewidth}
\vspace{-1.5cm}
\resourcerow
\vspace{6pt}
\begin{abstract}
\vspace{-1em}
Human visual reasoning is driven by systematic exploration and extraction of task-relevant visual evidence. Human gaze provides a window into this process, making gaze a useful source of guidance for visually-grounded reasoning in vision-language models (VLMs). However, human gaze also includes noise and redundancies that may transfer poorly to VLMs. In this work, we investigate how human gaze data can be effectively leveraged to improve visual reasoning in VLMs, while maintaining alignment with human behavior. We collect 7,109 human gaze trials from a diverse and challenging set of visual reasoning tasks, and train grounded reasoning models with different forms of gaze guidance. Across evaluations in chart understanding, visual search, and counting, we find that using human gaze as a prior for grounded reasoning consistently outperforms supervision using the full gaze sequence. Based on these findings, we introduce GGR (Gaze-Guided Reasoner), which achieves greater human alignment and more reliable visual grounding while matching or outperforming leading grounded-reasoning baselines in accuracy. Reinforcement learning further improves task accuracy, gaze alignment, and grounding quality despite using no gaze supervision. Behavioral analyses show that gaze supervised models exhibit more visual exploration before arriving at an answer. Our results suggest that the value of human gaze lies in guiding spatial exploration and shaping reasoning behavior without requiring complete imitation of human behavior.
\end{abstract}
\vspace{4pt}
\begingroup
\captionsetup{hypcap=false}
\providecommand{\teasergraphicwidth}{0.97\linewidth}
\newsavebox\teaserbox
\savebox\teaserbox{\begin{minipage}{\textwidth}
    \centering
    \setlength{\parskip}{0pt}
    \includegraphics[width=\teasergraphicwidth]
    {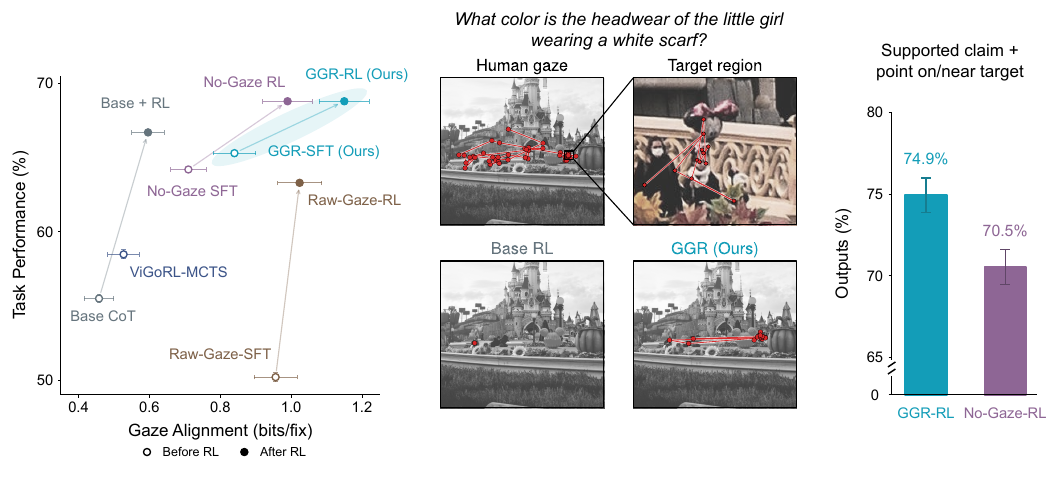}\par
    \begingroup
    \small
    \let\normalsize\small
    \setlength{\abovecaptionskip}{4pt}
    \captionof{figure}{\textbf{\model{} achieves higher human gaze alignment and better visual grounding while maintaining strong task performance.} (Left) Mean accuracy versus human fixation log-likelihood relative to uniform ($\pm$ SEM); open/filled circles denote pre/post-RL models.
    (Middle) Human gaze, Base RL, and \model{} grounding traces on a held-out example.
    (Right) Supported grounded reasoning claims with points on/near target for GGR-RL and No-Gaze-RL ($\pm$ SEM). All results on Qwen3-VL-8B-Instruct.}
    \label{fig:teaser}
    \endgroup
\end{minipage}}
\typeout{TEASER: needs \the\dimexpr\ht\teaserbox+\dp\teaserbox\relax; free on this page \the\dimexpr\pagegoal-\pagetotal\relax}
\par\noindent\usebox\teaserbox\par

\endgroup
\clearpage

\section{Introduction}

Visual reasoning is fundamentally a process of \emph{selective exploration}. A reasoner must decide where to look, ground abstract concepts to locations in the visual world, and use that evidence to determine where to look next. Humans perform this visual exploration via their gaze. When comparing two series in a chart, for example, a person may look from the legend to one curve, trace it to a relevant point, inspect the corresponding axis value, and then repeat the process for the other curve. When searching for a person in a crowded scene, humans move their eyes to different people in the scene to check for identifying attributes and reject distracting content. Human gaze therefore provides a naturally grounded record of task-directed visual exploration.

This signal is particularly appealing for training vision-language models (VLMs). Although chain-of-thought reasoning allows VLMs to decompose a problem into textual steps, those steps may not remain faithfully connected to the image. Models may confuse which features belong to which objects \citep{campbell2025understandinglimitsvisionlanguage,budny2026visualserialprocessingdeficits}, or hallucinate visual evidence, producing an internally coherent trace whose intermediate claims are unsupported by the image \citep{liu2025morethinking,ma2026understanding}. Grounded reasoning methods address such failures by associating reasoning steps with explicit image locations, forcing models to cite evidence they are referring to in text \citep{bindingpointbypoint}. Prior methods have explored training this grounded reasoning using tree search distillation or on-policy reinforcement learning (RL) to discover sequences of spatially grounded reasoning steps~\citep{sarch2026groundedreinforcementlearningvisual,cao2026groundr1incentivizinggroundedvisual}. These approaches can discover useful visual reasoning strategies, but depend on the model itself to ground reasoning steps accurately and propose exploration locations.

Human behavior offers a complementary source of guidance. Progress in language-model reasoning has benefited from human-written solutions and step-level annotations~\citep{cobbe2021gsm8k,lightman2023letsverify}. Beyond answer correctness, the behaviors demonstrated in chain-of-thought---such as verification and backtracking---can shape subsequent improvement through RL~\citep{gandhi2025}. For visual reasoning, gaze records precise spatial regions that people look at while exploring an image and processing visual evidence. Could gaze recordings provide similarly rich supervision for grounded reasoning?

In this work, we investigate which forms of gaze supervision support both strong visual reasoning and closer alignment with human evidence selection. Evaluating both outcomes is important to understand under what conditions gaze guidance transfers to VLMs. Task accuracy does not reveal whether models spatially explore similarly to humans, while gaze similarity alone does not establish that the transferred behavior supports effective reasoning. Human fixations also reflect perceptual and memory constraints that VLMs do not share~\citep{hayhoe2005eye,ballard1997deictic}, so reproducing them in full may not improve both aims. 

We systematically study how different constraints on gaze supervision transfer to VLMs in terms of task performance and human alignment. We collect 7,109 eye-tracking trials while human participants perform challenging chart understanding, counting, and visual search tasks. For each collected trial, we then provide an annotation model with the set of human fixations, along with the image and question, and have it annotate grounded reasoning under various constraints on the use of the provided human gaze (e.g., annotate every fixation or a selected subset of fixations). We finetune a compact model on the constructed datasets and evaluate them on challenging visual reasoning benchmarks. Our findings demonstrate that selective use of gaze is more effective than supervising every fixation for jointly improving both gaze alignment and task performance. 

Based on our findings, we introduce \model{} (Gaze-Guided Reasoner), which uses human fixations as a spatial prior for constructing grounded reasoning supervision. After gaze-guided supervised finetuning, \modelsft{} obtains the highest task performance and gaze alignment relative to all grounded baselines considered. After RL on \modelsft{} without gaze supervision, gaze-guided trained models further improve on human-gaze alignment, obtaining higher alignment than the strongest baselines while maintaining competitive task performance. Beyond these gains in gaze alignment, \model{} produces reasoning whose visual claims correspond more reliably to the reported image locations compared to no-gaze baselines. \modelrl{} also exhibits a distinct reasoning behavioral profile, expressing more initial exploration and reference-based navigation than no-gaze baselines. These results point to human gaze as a promising and largely untapped source of guidance for building AI systems whose visual reasoning is both effective and accurately grounded in the evidence people use.

\clearpage
\section{Related work}
\label{sec:related}

\textbf{Grounded reasoning in VLMs.}
\label{sec:related_grounded-VLMs}
Difficulties in binding object features in multi-object scenes have motivated recent methods that train VLMs to associate their reasoning steps with explicit image locations \citep{campbell2025understandinglimitsvisionlanguage,budny2026visualserialprocessingdeficits}. Point-based supervision can improve object-centric processing \citep{deitke2024molmopixmoopenweights,bindingpointbypoint}, and points or bounding boxes can be incorporated directly into reasoning traces to improve performance across broader grounded reasoning tasks. Such supervision is straightforward when the task specifies what to identify, but more complex multi-step tasks require discovering a sequence of relevant regions. ViGoRL-MCTS constructs grounded traces through tree search before applying RL \citep{sarch2026groundedreinforcementlearningvisual}, while related methods use answer-based rewards to learn where to gather visual evidence, including which regions to crop or zoom into \citep{cao2026groundr1incentivizinggroundedvisual,wang2025pixelreasonerincentivizingpixelspace}. These approaches support flexible grounding, although the strategies available for training depend on model-generated proposals. We use human gaze to guide grounded reasoning trace proposal, comparing against ViGoRL-MCTS to assess the relative benefit of human- vs. model-guided traces.

\textbf{Eye movements and visual reasoning.}
\label{sec:related_human-gaze}
Eye movements constrain how people gather visual information in response to task demands. Different questions about the same image elicit different fixation patterns \citep{yarbus1967eye}, and during natural behavior, gaze is directed toward locations relevant to the current reasoning step, often just before the information is needed \citep{hayhoe2005eye,land2001ways}. Fixation sequences also provide evidence about the coordination of perceptual and cognitive operations \citep{just1976eye}, including the interplay between reading and interpreting chart elements \citep{carpenter1998model} and the shifts in attention associated with problem solving \citep{grant2003eye}. Repeated fixations can allow people to consult external information instead of retaining it in memory \citep{ballard1995memory}. \citet{ballard1997deictic} describe such fixations as pointers connecting cognitive operations to locations in the world, emphasizing how eye movements help organize the use of visual evidence over time.

\textbf{Learning from human behavior.}
\label{sec:related_gaze-supervision}
Human behavioral signals have been used to train neural networks toward closer correspondence with human processing \citep{linsley2019learning,fel2022harmonizing,muttenthaler2025aligning,sucholutsky2025getting,binz2025centaur,mondal2026gaze}. Such correspondence is not captured by accuracy alone, since comparable performance can conceal different solution strategies \citep{geirhos2020beyond,matchedoutcomes}. Human gaze has been used to supervise attention to task-relevant information. One line of work aggregates fixations into spatial salience maps, supervising where a model attends without retaining the order \citep{qiao2017exploringhumanlikeattentionsupervision,saran2021efficientlyguidingimitationlearning,salamatian2025chartgazeenhancingchartunderstanding}. A second line retains ordered fixation trajectories to supervise sequential visual processing, as explored in recent radiology models \citep{li2026thinking,lee2026seeingexpertseyesfoundational,zhu2026medeyeslearningdynamicvisual}. However, agreement with human attention does not consistently predict task accuracy \citep{Sood_2021}, and fixation sequences vary across observers and reflect constraints that VLMs do not share \citep{wilming2011limits,matchedoutcomes}. Our work examines gaze as selective \textit{guidance} for reasoning-trace generation, evaluating task performance and behavioral alignment separately.
\vspace{-8pt}
\section{Method}
\label{sec:prelims}

\subsection{Problem setup}
Our objective is to learn a vision-language model (VLM) that obtains strong visual reasoning performance while selecting visual evidence aligned with human gaze behavior.
We consider a training dataset $\mathcal{D} = \{(I_i,q_i,a_i^\ast,F_i)\}_{i=1}^{N},$ where $I_i$ is an image, $q_i$ is a natural-language question, $a_i^\ast$ is the ground-truth answer, and $F_i=(f_{i,1},\ldots,f_{i,M_i})$ is a human fixation trace recorded during problem solving. Each $f_{i,j}=(x_{i,j},y_{i,j})$ denotes a fixation location in image coordinates. Additional details of human data collection can be found in Section~\ref{sec:data_collection}.

We represent a grounded reasoning step as $n_t=\langle s_t,p_t\rangle$, pairing a textual thought $s_t$ with an image location $p_t$ that identifies the visual evidence it refers to. A trajectory $\tau=(n_1,\ldots,n_T,a)$ consists of these steps followed by a predicted answer $a$. Our goal is to design a learning algorithm $\mathcal{A}$ that maps the training dataset $\mathcal{D}$ to a grounded reasoning policy $\pi_\theta=\mathcal{A}(\mathcal{D})$, with the objective of improving answer accuracy and gaze alignment on unseen examples. At inference time, $\pi_\theta$ conditions only on the image and question $(I,q)$. Human fixations are available only during training.

\begin{figure}[!htbp]
    \centering
    \includegraphics[width=\linewidth]{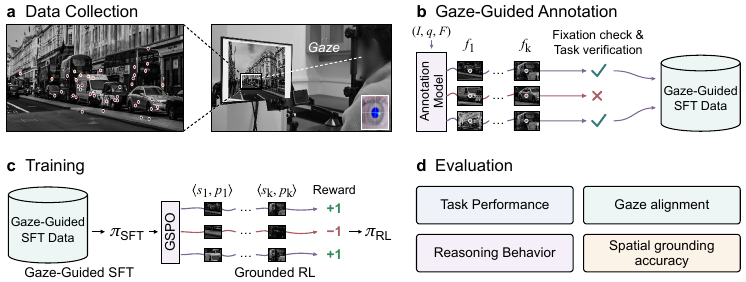}
    \caption{\textbf{Learning grounded reasoning with gaze guidance.}
    \textbf{(a)} Eye gaze was recorded while human participants solved visual reasoning problems.
    \textbf{(b)} We give the image $I$, question $q$, and fixation list $F$ to an annotation model, which annotates regions guided by human fixations $f_k$ to form a grounded reasoning trace.
    \textbf{(c)} We train on these traces using SFT followed by RL, where each grounded step $\langle s_k,p_k\rangle$ pairs a reasoning statement $s_k$ with its grounding location $p_k$.
    \textbf{(d)} We evaluate the models on task performance, gaze alignment, reasoning behavior, and spatial grounding accuracy.
    }
    \label{fig:methods}
\end{figure}

\subsection{Learning Grounded Reasoning from Human Gaze.}

Our goal is to learn a grounded reasoning trajectory $\tau=(\langle s_1,p_1\rangle,\ldots,\langle s_T,p_T\rangle,a)$, where $p_t$ is the \emph{evidence proposal} and $s_t$ is the \emph{reasoning supervision}. 
While human fixation traces contain useful task-directed information, they may not themselves be ideal supervision as the \emph{evidence proposal} for a VLM. A fixation may correspond to identifying a target, inspecting an attribute, rejecting a distractor, verifying an intermediate conclusion, or simply revisiting previously observed evidence. Moreover, human gaze paths reflect perceptual and memory constraints that do not apply to VLMs. 

Creating a strong visual reasoning trace therefore requires deciding which fixation locations should be associated with a reasoning step, and whether to allow grounding at additional locations. We investigate how these choices affect task performance and human gaze alignment. To do so, we use an annotation model (Qwen3.5-122B-A10B) to construct grounded reasoning traces under different constraints on the use of human gaze. We detail this process below.

\subsubsection{Constructing Gaze-Guided Reasoning Supervision}\label{sec:reasoning_annotation}

We provide a strong annotation model (Qwen3.5-122B-A10B) with the image $I$, question $q$, and fixation coordinate list $F$, and prompt it to generate a grounded reasoning trace $\tau=(\langle s_t,p_t\rangle,\ldots,n_T,a)$, where $(p_1, \ldots, p_T)$ is guided by $F$ (Figure~\ref{fig:methods}b). We investigate different forms of gaze guidance in Section~\ref{sec:gaze-supervision-strategies}. These include constraints on whether each fixation must be annotated with a reasoning step, and if the annotation model can add additional regions outside the human fixations.
Importantly, we do not provide the reference answer $a^\ast$ while the annotation model constructs the reasoning trace. The model must derive an answer from the image and selected evidence rather than generate a post-hoc rationale for the ground-truth answer~\citep{zelikman2022star}. 

\textbf{Sampling and verification.}
For each trial, we sample 25 candidate trajectories. We apply rejection sampling~\citep{touvron2023llama}, giving the image, reasoning trace, and answer back to the annotation model to have it flag samples for 1) wrong answers, 2) redundancy in the model's response, 3) inconsistency between thinking and final answer, and 4) self-contradictions. We remove any sample from training which triggers any of these flags. The accepted trajectories form our gaze-guided SFT dataset. Annotation prompts and details are provided in Appendix~\ref{sec:app_annotation}.

\subsubsection{Training the Gaze-Guided Grounded Reasoner}

After obtaining the gaze-guided grounded reasoning supervision from the process in Section~\ref{sec:reasoning_annotation}, we train a base VLM to mimic the grounded traces, followed by reinforcement learning using task accuracy rewards without gaze guidance.
Our training pipeline is shown in Figure~\ref{fig:methods}c. We detail each stage below.

\textbf{Supervised fine-tuning.}
Using the fixation-guided dataset we obtain from Section~\ref{sec:reasoning_annotation}, we train a compact model with supervised finetuning on the dataset to encourage grounding aligned with gaze behavior. The resulting model thus learns to mimic the gaze-guided reasoning in the training traces, using only the image and question at inference time.

\textbf{Reinforcement Learning.}
We then apply on-policy reinforcement learning to \modelsft{} to refine the grounded model behavior without human gaze. We follow the Vero~\citep{sarch2026vero} recipe, and use GSPO~\citep{zheng2025groupsequencepolicyoptimization} with clip-higher, no KL penalty, and soft overlong penalty. The total reward combines 95\% answer accuracy and 5\% format compliance (\texttt{\textless think\textgreater} tags, boxed answer, and at least one valid coordinate). Our accuracy reward uses text matching for short answers, math-verify for numerical answers, and absolute error decreasing exponentially for counting rewards. 

\textbf{Implementation details.} For supervised finetuning of our models and baselines, we train five epochs and choose the best checkpoint as determined by the validation accuracy. We use the 698 human trials as our validation set and only select based on accuracy. For reinforcement learning, all models train 400 steps, taking the best checkpoint by validation accuracy. For RL, we use 100,000 Vero chart QA examples~\citep{sarch2026vero}, 9,667 visual search examples from PixelReasoner and Visual Probe subsets in Vero~\citep{sarch2026vero}, and 7,785 counting examples curated from FSC147, REC-8K, PixMo, and RPINE. We sample batches with equal representation from each domain. We provide SFT and RL training configurations in Tables~\ref{tab:sft_setup} and~\ref{tab:rl_setup}, respectively.
Full SFT and RL hyperparameters are provided in Appendix~\ref{sec:app_train_eval_details}.

\subsubsection{Collecting Human Gaze Traces}
\label{sec:data_collection}
To collect human gaze traces for reasoning annotation, we recruited participants to solve visual reasoning problems while we tracked their eye gaze (Figure~\ref{fig:methods}a). Participants completed chart question answering, counting, and visual search trials while participating in the study. Images were scaled so that their longer dimension occupied 1,267 pixels, with questions presented below the image. Real-time synchronization between PsychoPy and the eye tracker allowed participants to zoom into or out of the image around their current gaze location. Participants could also place markers at their gaze location to keep track of counted objects or previously searched regions, reducing working-memory demands during extended trials. We collected 7,109 trials (2,145 visual search, 2,652 counting, 2,312 chart) in total with fixation coordinate traces for each trial. We use 6,411 trials for SFT training and 698 for held-out evaluation trials. Additional details on data collection are in Appendix~\ref{sec:app_human_data_coll}.

\section{Results}
\definecolor{resultshighlight}{HTML}{E6EEF8}

Our experiments aim to understand how human gaze supervision affects visual reasoning performance and gaze alignment in vision-language models. As shown in Figure~\ref{fig:methods}d, we seek to understand how each condition and baseline performs on task accuracy, gaze alignment, spatial grounding accuracy, and chain of thought behavior.

\label{sec:evaluation}
\textbf{Visual reasoning evaluation suite.} We curate a suite of challenging and diverse benchmarks to test the ability of our method and baselines on novel problems in chart understanding, counting, and visual search. We evaluate on ChartQA~\citep{masry2022chartqabenchmarkquestionanswering} and ChartQA-Pro~\citep{masry2025chartqaprodiversechallengingbenchmark} for chart understanding, CountBenchQA~\citep{paiss2023teachingclipcount} and CountQA~\citep{tamarapalli2025countqamllmscountwild} for counting, and MME-RealWorld-Lite~\citep{zhang2025mmerealworldmultimodalllmchallenge} and Visual Probe~\citep{lai2025minio3scalingreasoningpatterns} for visual search in high resolution images. 

\textbf{Gaze alignment.}
We measure spatial correspondence between model grounding points and human fixations. We form a Gaussian kernel density estimate over model coordinates with standard deviation $\sigma$ and uniform background weight $\epsilon$, then compute the mean log-likelihood of human fixations relative to that of a uniform spatial baseline (following \citet{deepgaze}). For each model tested, we fit $(\sigma_c,\epsilon_c)$ via grid search. Of the 698 held-out trials, 336 are used to calibrate the gaze-alignment metric and 362 to evaluate it. Behavioral analyses use all 698 held-out trials. The metric does not capture fixation order, duration, or the model's internal attention. See Appendix~\ref{sec:app_fix_align} for additional details.

\definecolor{resultshighlight}{HTML}{E6EEF8} 
\definecolor{resultscolumnhighlight}{HTML}{F2F6FB} 
\begin{table}[!htbp]
    \centering
    \begingroup
    \setlength{\tabcolsep}{3pt}
    \renewcommand{\arraystretch}{1.15}
    \resizebox{\textwidth}{!}{%
    \begin{tabular}{@{}l*{3}{c}>{\columncolor{resultscolumnhighlight}}c>{\columncolor{resultscolumnhighlight}}c@{\hspace{9pt}}*{3}{c}>{\columncolor{resultscolumnhighlight}}c>{\columncolor{resultscolumnhighlight}[\tabcolsep][0pt]}c@{}}
    \toprule
    & \multicolumn{5}{c}{\textbf{SFT}} & \multicolumn{5}{c}{\textbf{SFT $\rightarrow$ RL}} \\
    \cmidrule(lr){2-6}\cmidrule(lr){7-11}
    \textbf{Condition} & \textbf{Chart} & \textbf{Counting} & \textbf{Search} & \textbf{Overall} & \shortstack{\textbf{Gaze}\\\textbf{alignment}\\{[bits/fix]}} & \textbf{Chart} & \textbf{Counting} & \textbf{Search} & \textbf{Overall} & \shortstack{\textbf{Gaze}\\\textbf{alignment}\\{[bits/fix]}} \\
    \midrule
    Human split-half & -- & -- & -- & -- & $1.97^{\pm 0.08}$ & -- & -- & -- & -- & $1.97^{\pm 0.08}$ \\
    \midrule
    \multicolumn{11}{@{}l}{\textit{Controls}} \\
    No-Gaze & $75.1^{\pm 0.0}$ & $67.7^{\pm 0.3}$ & $49.9^{\pm 0.5}$ & $64.2^{\pm 0.2}$ & $0.71^{\pm 0.05}$ & $77.8^{\pm 0.2}$ & $74.0^{\pm 0.1}$ & $54.6^{\pm 0.2}$ & $\mathbf{68.8}^{\pm 0.1}$ & $0.99^{\pm 0.07}$ \\
    Shuffled & $69.1^{\pm 0.2}$ & $51.4^{\pm 1.0}$ & $47.0^{\pm 0.5}$ & $55.8^{\pm 0.4}$ & $0.01^{\pm 0.00}$ & -- & -- & -- & -- & -- \\
    \midrule
    \multicolumn{11}{@{}l}{\textit{Human-gaze supervision}} \\
    Raw-Gaze & $53.6^{\pm 0.1}$ & $54.9^{\pm 0.6}$ & $41.9^{\pm 0.5}$ & $50.2^{\pm 0.3}$ & $0.96^{\pm 0.06}$ & $75.3^{\pm 0.2}$ & $60.6^{\pm 0.3}$ & $53.9^{\pm 0.2}$ & $63.3^{\pm 0.1}$ & $1.02^{\pm 0.06}$ \\
    Gaze-All & $69.3^{\pm 0.2}$ & $53.1^{\pm 0.2}$ & $45.3^{\pm 0.5}$ & $55.9^{\pm 0.2}$ & $\mathbf{1.04}^{\pm 0.06}$ & $75.4^{\pm 0.1}$ & $65.7^{\pm 0.2}$ & $54.8^{\pm 0.3}$ & $65.3^{\pm 0.1}$ & $1.03^{\pm 0.07}$ \\
    Gaze-Subset & $74.8^{\pm 0.1}$ & $63.1^{\pm 0.3}$ & $50.6^{\pm 0.6}$ & $62.8^{\pm 0.2}$ & $0.82^{\pm 0.05}$ & $76.6^{\pm 0.2}$ & $69.9^{\pm 0.2}$ & $55.4^{\pm 0.3}$ & $67.3^{\pm 0.1}$ & $0.99^{\pm 0.07}$ \\
    \model{} & $74.6^{\pm 0.3}$ & $68.4^{\pm 0.2}$ & $52.9^{\pm 0.4}$ & $\mathbf{65.3}^{\pm 0.2}$ & $0.84^{\pm 0.06}$ & $78.0^{\pm 0.4}$ & $73.3^{\pm 0.2}$ & $55.2^{\pm 0.3}$ & $\mathbf{68.8}^{\pm 0.2}$ & $\mathbf{1.15}^{\pm 0.07}$ \\
    \bottomrule
    \end{tabular}%
    }
    \endgroup
    \caption{\textbf{Comparing strategies for human-gaze supervision.} Task accuracy (\%) and human-gaze alignment after SFT and subsequent RL on Qwen3-VL-8B-Instruct. Accuracy averages the two benchmarks in each domain or all six overall. Reported is Avg@3 ($\pm$ SEM) across three evaluation runs and gaze alignment is mean $\pm$ SEM in bits/fix. Benchmark-level results are in Appendix Table~\ref{tab:sft-rl-task-results}.}
    \label{tab:gaze-supervision-strategies}
\end{table}

\subsection{Human gaze supervision is most effective when used as a flexible prior}
\label{sec:gaze-supervision-strategies}

To systematically study the impact of different forms of gaze-guided supervision on task performance and human alignment, we train on data generated by conditioning a strong model (Qwen3.5-122B-A10B) on gaze while varying the constraints as outlined in Section~\ref{sec:prelims}. In particular, we compare the following conditions: 
(1) the annotation model annotates each fixation with a textual reasoning step (\emph{Gaze-All}), (2) the annotation model selects a strict subset of the human fixations and annotates each selected fixation with a reasoning step (\emph{Gaze-Subset}), (3) the annotation model uses the fixation list as a prior, choosing which fixations to annotate and grounding additional regions when needed (Gaze Guided Reasoner; \model{}), (4) we directly use fixation coordinates interleaved with transcribed human think aloud (\emph{Raw-Gaze}), (5) we use the Gaze-All condition with fixation lists shuffled across examples (\emph{Shuffled}), (6) the annotation model generates grounded reasoning without human fixation guidance (\emph{No-Gaze}). We report the results in Table~\ref{tab:gaze-supervision-strategies}.

We investigate which constraints on human gaze supervision lead to improvements in task performance and gaze alignment. We first compare Gaze-All, where we require a textual explanation at every fixation, with Gaze-Subset, where we allow the annotation model to select which fixations to include. After SFT on Qwen3-VL-8B-Instruct, overall accuracy increases from $55.9\%$ with Gaze-All to $62.8\%$ with Gaze-Subset (Table~\ref{tab:gaze-supervision-strategies}). We observe improvements on all six benchmarks (Appendix Table~\ref{tab:sft-rl-task-results}). Although gaze alignment is higher with Gaze-All than with Gaze-Subset ($1.04$ versus $0.82$ bits/fix), task accuracy is lower. Gaze-All exhibits significant re-examination behavior (increase of $54.8$ percentage points relative to the base model after RL; $p<0.001$; Figure~\ref{fig:behavioral-contrasts}a), which often leads to self-confusion and repeated loops. These results suggest that selective use of gaze is more effective than supervising every fixation.

\begin{findingbox}
Selective use of gaze is more effective than supervising every fixation.
\end{findingbox}

We next ask whether the selected evidence should be restricted only to locations that humans fixated. With Gaze-Subset, fixations can be omitted but cannot include evidence from other regions. This restriction may exclude evidence that the model finds useful, since a person may not look directly at every region that could support the answer. To test whether this additional flexibility improves reasoning, we use gaze as a prior in \model{} (Gaze-Guided Reasoner). We treat human fixations as a source of \emph{candidate visual evidence} by providing the annotation model with the fixation list, while allowing the model to select which regions to include in a grounded reasoning trace, and add additional regions when needed.

After SFT, overall accuracy is $2.5$ percentage points higher with \modelsft{} than with Gaze-Subset. After RL without gaze supervision, overall accuracy remains $1.5$ percentage points higher with \modelrl{} than with Gaze-Subset, while improving gaze alignment ($1.15$ \modelrl{} versus $0.99$ Gaze-Subset bits/fix; $p<0.001$). Both conditions remain more accurate overall than Gaze-All (Table~\ref{tab:gaze-supervision-strategies}). We observe the same ordering after SFT with Gemma 3 4B and MiMo-VL-7B, with overall accuracy highest for \modelsft{}, followed by Gaze-Subset and Gaze-All (Appendix Tables~\ref{tab:gemma3-4b-sft-task-results} and~\ref{tab:mimo-sft-task-results}). We thus find using gaze as a prior (\model{}) improves task accuracy and gaze alignment most among conditions tested.

\begin{findingbox}
Selective use of gaze is more effectively implemented as a prior rather than as a strict subset.
\end{findingbox}

Our results suggest that better reasoning and greater human-gaze alignment can coexist, but stronger alignment alone does not imply better performance.
After SFT, gaze alignment is higher with Gaze-All than with \modelsft{} ($1.04$ versus $0.84$ bits/fix; $p<0.001$), although overall accuracy is lower. We observe the same pattern with Raw-Gaze-SFT (Table~\ref{tab:gaze-supervision-strategies}). After removing the correspondence between gaze and the task by shuffling fixation traces across examples, gaze alignment falls to $0.01$ bits/fix, while overall accuracy remains similar ($55.8\%$ versus $55.9\%$ for Gaze-All). We therefore find that task-specific gaze guidance matters for human alignment even when accuracy changes little.

\begin{table*}[!htbp]
    \centering
    \begingroup
    \setlength{\tabcolsep}{6pt}
    \renewcommand{\arraystretch}{1.15}
    \resizebox{\textwidth}{!}{    \begin{tabular}{@{}l*{6}{c}>{\columncolor{resultscolumnhighlight}}c>{\columncolor{resultscolumnhighlight}[\tabcolsep][0pt]}c@{}}
    \hline
    & \multicolumn{2}{c}{\textbf{Chart}} & \multicolumn{2}{c}{\textbf{Counting}} & \multicolumn{2}{c}{\textbf{Search}} & \multicolumn{2}{c@{}}{} \\
    \cmidrule(lr{3pt}){2-3}\cmidrule(lr{3pt}){4-5}\cmidrule(lr{3pt}){6-7}
    \noalign{\vskip 3pt}
    \textbf{Condition} & \shortstack{ChartQA\\Pro} & ChartQA & \shortstack{Count\\BenchQA} & CountQA & \shortstack{MME\\RealWorld\\Lite} & \shortstack{Visual\\Probe} & \textbf{Overall} & \shortstack{\textbf{Gaze}\\\textbf{Alignment}\\{[bits/fix]}} \\
    \hline
    Human split-half & -- & -- & -- & -- & -- & -- & -- & $1.97^{\pm 0.08}$ \\
    \hline
    \multicolumn{9}{@{}l}{\textit{Reference models}} \\
    GPT-5-Nano & $56.0$ & $80.1$ & $75.4$ & $25.7$ & $49.8$ & $41.5$ & $54.8$ & -- \\
    Molmo2-O-7B & $31.4$ & $75.2$ & $89.4$ & $32.1$ & $44.4$ & $34.8$ & $51.2$ & -- \\
    DeepEyes-7B & $36.3$ & $76.7$ & $80.0$ & $18.7$ & $49.0$ & $40.6$ & $50.2$ & -- \\
    Vero-Qwen3I-8B & $63.0$ & $91.9$ & $92.7$ & $33.4$ & $57.2$ & $47.0$ & $64.2$ & -- \\
    Qwen3-VL-8B-Instruct & $46.0^{\pm 0.2}$ & $82.6^{\pm 0.2}$ & $86.8^{\pm 1.3}$ & $28.4^{\pm 0.5}$ & $44.7^{\pm 0.3}$ & $44.5^{\pm 0.6}$ & $55.5^{\pm 0.3}$ & $0.46^{\pm 0.04}$ \\
    \hline
    \noalign{\vskip 1pt}
    \multicolumn{9}{@{}l@{}}{\textit{Supervised fine-tuning (SFT)}} \\
    \noalign{\vskip 2pt}
    ViGoRL-MCTS & $46.2^{\pm 0.8}$ & $84.8^{\pm 0.4}$ & $90.4^{\pm 0.5}$ & $30.0^{\pm 0.4}$ & $53.6^{\pm 0.4}$ & $45.7^{\pm 0.6}$ & $58.4^{\pm 0.2}$ & $0.53^{\pm 0.04}$ \\
    No-Gaze & $59.6^{\pm 0.1}$ & $90.5^{\pm 0.0}$ & $94.9^{\pm 0.5}$ & $40.5^{\pm 0.3}$ & $54.2^{\pm 0.2}$ & $45.6^{\pm 0.9}$ & $64.2^{\pm 0.2}$ & $0.71^{\pm 0.05}$ \\
    \modelsft{} & $58.4^{\pm 0.5}$ & $90.8^{\pm 0.0}$ & $95.0^{\pm 0.4}$ & $41.8^{\pm 0.3}$ & $57.7^{\pm 0.1}$ & $48.0^{\pm 0.7}$ & $\mathbf{65.3}^{\pm 0.2}$ & $\mathbf{0.84}^{\pm 0.06}$ \\
    \hline
    \noalign{\vskip 1pt}
    \multicolumn{9}{@{}l@{}}{\textit{Reinforcement learning (RL)}} \\
    \noalign{\vskip 2pt}
    Point-Grounded RL & $57.0^{\pm 0.4}$ & $91.4^{\pm 0.0}$ & $96.1^{\pm 0.4}$ & $44.4^{\pm 0.4}$ & $59.2^{\pm 0.5}$ & $52.2^{\pm 0.7}$ & $66.7^{\pm 0.2}$ & $0.60^{\pm 0.05}$ \\
    Ungrounded RL & $59.1^{\pm 0.3}$ & $92.3^{\pm 0.0}$ & $92.7^{\pm 0.4}$ & $39.1^{\pm 0.3}$ & $57.6^{\pm 0.2}$ & $53.1^{\pm 0.4}$ & $65.7^{\pm 0.1}$ & -- \\
    No-Gaze-SFT $\rightarrow$ RL & $63.6^{\pm 0.4}$ & $92.0^{\pm 0.1}$ & $95.1^{\pm 0.0}$ & $52.9^{\pm 0.1}$ & $57.3^{\pm 0.3}$ & $51.8^{\pm 0.2}$ & $\mathbf{68.8}^{\pm 0.1}$ & $0.99^{\pm 0.07}$ \\
    \modelrl{} & $64.0^{\pm 0.7}$ & $91.9^{\pm 0.3}$ & $95.9^{\pm 0.1}$ & $50.7^{\pm 0.5}$ & $59.6^{\pm 0.2}$ & $50.8^{\pm 0.6}$ & $\mathbf{68.8}^{\pm 0.2}$ & $\mathbf{1.15}^{\pm 0.07}$ \\
    \hline
    \end{tabular}    }
    \endgroup
    \caption{\textbf{GGR compared with existing models and training baselines.} Task performance and human-gaze alignment for reference models, SFT methods, and RL methods. All SFT and RL methods use Qwen3-VL-8B-Instruct. Task scores report Avg@3 ($\pm$ SEM) across three evaluation runs. Overall is the unweighted mean of the six benchmarks. Gaze alignment is mean $\pm$ SEM in bits/fix.}
    \vspace{-6pt}
    \label{tab:ggr-baselines}
\end{table*}

\subsection{\model{} improves gaze alignment and reasoning quality without sacrificing performance}
\label{sec:ggr-baselines}

Building on the findings of the previous section, we now examine how our best gaze-guided method, \model{}, compares to grounded reasoning baselines that do not use gaze guidance. We compare with ViGoRL-MCTS~\citep{sarch2026groundedreinforcementlearningvisual}, for which we generate the training traces through tree search, No-Gaze, where we use the same annotation model and procedure as \model{} to generate grounded reasoning without providing human fixations, and several open and closed general reasoning models~\citep{clark2026molmo2openweightsdata,singh2025openai,qwen35blog} (Table~\ref{tab:ggr-baselines}).

After SFT, \model{} obtains the highest overall accuracy among the models considered, exceeding the strongest baseline, No-Gaze, by $1.1$ percentage points in overall performance and $0.13$ bits/fix in gaze alignment ($p<0.001$). After RL, \modelrl{} improves overall accuracy from $65.3\%$ to $68.8\%$ and gaze alignment from $0.84$ to $1.15$ bits/fix ($p<0.001$), matching No-Gaze in performance while achieving higher gaze alignment ($1.15$ versus $0.99$ bits/fix; $p<0.001$). \model{} therefore maintains the accuracy of the strongest baseline while grounded reasoning is more aligned with human gaze.

\begin{findingbox}
GGR maintains the accuracy of the strongest baseline while being more aligned with human gaze.
\end{findingbox}

Given that gaze provides precise candidate locations for grounding as humans reason, we further investigate whether gaze-guided supervision improves the spatial grounding and reasoning quality. We use GPT-6.1-Sol-xhigh to assess grounding locations and their associated reasoning steps ($N=1{,}500$). We validate the model ratings with two human raters ($N=100$ samples; $87$--$90\%$ agreement with model; Appendix Table~\ref{tab:human-grounding-agreement}). 
Full methods are in Appendix~\ref{sec:app_add_results}.

\begin{figure}[!t]
    \centering
    \ifdefined\includevisualgroundingaudit
    \begin{minipage}[t]{0.27\linewidth}
        \vspace{0pt}
        \vspace*{19pt}
        \centering
        \includegraphics[width=\linewidth,trim=0bp 0bp 0bp 30bp,clip]{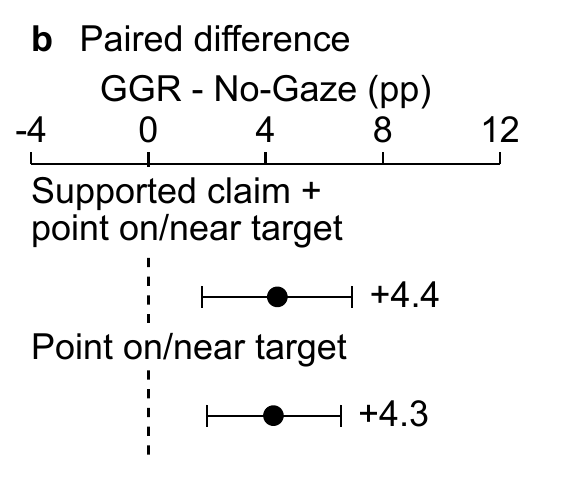}
    \caption{\textbf{GGR improves visual grounding quality compared to the no gaze condition.} Paired GGR minus No-Gaze differences in visual grounding after RL. Reported is mean $\pm$ bootstrapped 95\% CI. SFT results are in Appendix Figure~\ref{fig:visual-grounding-sft}.}
    \label{fig:visual-grounding}
    \end{minipage}\hfill
    \begin{minipage}[t]{0.71\linewidth}
        \vspace{0pt}
        \centering
    \fi
    \includegraphics[width=\linewidth]{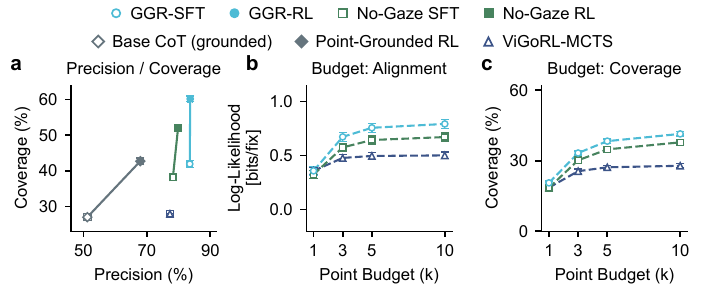}
    \caption{\textbf{Gaze alignment and coverage.}
    (a) Coverage versus precision at a radius of 100 normalized coordinate units.
    (b,c) Alignment and coverage at point budgets $k\in\{1,3,5,10\}$ for \modelsft{}, No-Gaze SFT, and ViGoRL-MCTS.
    Open and filled markers denote SFT and RL, respectively.
    Reported is mean $\pm$ SEM.}
    \label{fig:gaze-alignment-overview}
    \label{fig:accuracy-alignment}
    \label{fig:fixed-budget}
    \ifdefined\includevisualgroundingaudit
    \end{minipage}
    \fi
\end{figure}

Compared to our method without gaze guidance (No-Gaze), \model{} obtains a higher rate of supported claims that are correctly grounded after both SFT and RL, with gains of $4.5\%$ and $4.4\%$, respectively (Figure~\ref{fig:visual-grounding}; Appendix Figure~\ref{fig:visual-grounding-sft}; $p=0.0025$). 
\modelrl{} additionally achieves better grounding accuracy regardless of whether the claim is supported (Table~\ref{tab:visual-grounding-rates} \emph{Point on/near target}; $84.1\%$ with \modelrl{} compared with $79.8\%$ for No-Gaze-RL; $p=0.0016$). Our results demonstrate that gaze guidance improves the rate of supported reasoning claims and accuracy of the visual grounding.
\vspace{-0pt}
\begin{findingbox}
GGR produces visual reasoning claims that correspond more reliably to the reported image locations.
\end{findingbox}

\subsection{Gaze-guided supervision improves alignment coverage and precision}
\label{sec:gaze-alignment}

In section~\ref{sec:ggr-baselines}, we established that \model{} improves gaze alignment compared to the no gaze conditions. In this section, we investigate in what ways gaze-guided supervision improves this gaze alignment. We examine gaze coverage and precision between the model and human locations. We define gaze coverage as the fraction of human fixations within a fixed radius of a model grounding location and gaze precision as the fraction of unique model points that are within a fixed radius of a human fixation. We use 100 normalized coordinate units as the radius.

Compared to the initial base model, \modelsft{} significantly increases both coverage and precision (Figure~\ref{fig:gaze-alignment-overview}a; $41.9\%$ \modelsft{} vs. $27.0\%$ Base-CoT coverage; $83.5\%$ \modelsft{} vs. $51.2\%$ Base-CoT precision). Relative to ViGoRL-MCTS~\citep{sarch2026groundedreinforcementlearningvisual} and our method without gaze guidance (No-Gaze-SFT), \modelsft{} also improves both coverage and precision (Figure~\ref{fig:gaze-alignment-overview}a; all six comparisons $p<0.001$). We additionally consider a point-budget-matched coverage comparison, using the first $k$ points in the reasoning trace. Under this setup, \modelsft{} maintains improvements over both ViGoRL-MCTS and the No-Gaze-SFT condition (Figure~\ref{fig:gaze-alignment-overview}c). Thus, \modelsft{} improves both coverage and precision under full and budget matched comparisons to no-gaze baselines.

\begin{findingbox}
Relative to supervised baselines not using gaze, \modelsft{} increases both gaze coverage and precision.
\end{findingbox}

After applying RL without the use of gaze supervision, \modelrl{} remains the model with the highest coverage and precision among those tested (Figure~\ref{fig:gaze-alignment-overview}a; full coverage $60.2\%$ and precision $83.7\%$, coverage at $k=10$ is $54.8\%$ \modelrl{} vs. $49.6\%$ No-Gaze-RL and $41.3\%$ Point-Grounded RL; Appendix Figure~\ref{fig:additional-rl-budgets}b). RL on \modelsft{} largely improves coverage ($41.9\%\rightarrow60.2\%$), while precision remains high ($83.5\%\rightarrow83.7\%$). The same trend in coverage and precision persists when doing RL on the No-Gaze-SFT (Figure~\ref{fig:gaze-alignment-overview}a; coverage $38.2\%\rightarrow52.0\%$; precision $78.2\%\rightarrow79.8\%$), while Point-Grounded RL improves both coverage and precision similarly (Figure~\ref{fig:gaze-alignment-overview}a). 

\begin{findingbox}
RL on \modelsft{} increases gaze coverage while maintaining precision.
\end{findingbox}

\subsection{Gaze-guided supervision changes reasoning behavior}\label{sec:behavior}

\begin{figure}[!t]
    \centering
    \includegraphics[width=\linewidth]{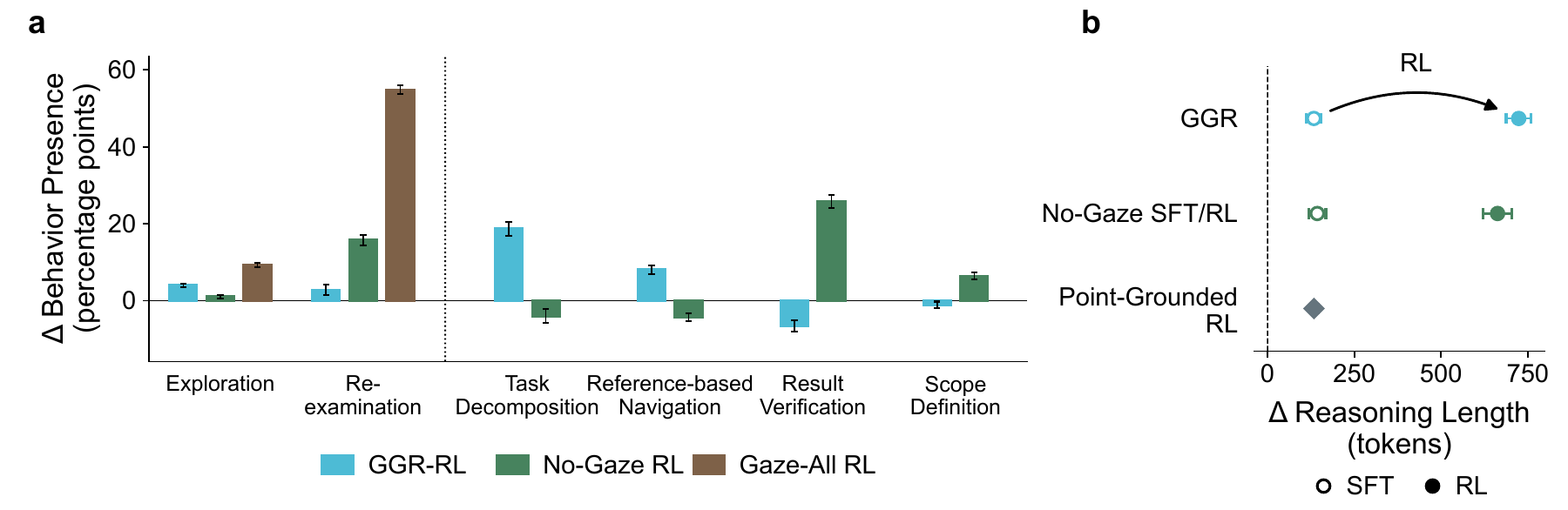}
    \caption{\textbf{Behavioral changes under gaze-guided and no-gaze training.} \textbf{a.} Changes relative to Qwen3-VL-8B-Instruct on six behaviors: visual exploration, re-examination, task decomposition, reference-based navigation, result verification, scope definition, in percentage points of behavior presence. \textbf{b.} Reasoning-length changes relative to the same base model in tokens. Open and filled markers denote selected SFT and RL checkpoints. Mean $\pm$ SEM ($N=698$).}
    \label{fig:behavioral-contrasts}
\end{figure}

To examine how gaze-guided training alters reasoning behavior, we analyze how the chains of thought behaviors of \modelrl{} and No-Gaze-RL change relative to the base model (Qwen3-VL-8B-Instruct) we are training. Following~\citet{gandhi2025}, for each behavior, we use a strong LLM (Qwen3.5-122B-A10B) to count occurrences in each chain of thought (held-out set; $N=698$). We report six behaviors which we determined based on inspecting the raw human fixation traces. These behaviors include visual exploration, re-examination, task decomposition, reference-based navigation, result verification, and scope definition. Full behavior definitions are provided in Appendix Table~\ref{tab:behavior_categories}. Presence denotes the average behavior presence across reasoning traces. We report the results in Figure~\ref{fig:behavioral-contrasts}.

We find that \modelrl{} increases exploration and task decomposition compared to the base model and the no gaze condition. After RL, \modelrl{} increases exploration by $3.9$ and task decomposition by $18.7$ percentage points of behavior presence relative to the base model, compared with $1.0$ and $-4.1$ for No-Gaze RL (Figure~\ref{fig:behavioral-contrasts}a). Reference-based navigation also increases ($+8.1$) compared with $-4.3$ for No-Gaze RL. Scope definition remains near the base model ($-1.1$) compared with $+6.4$ for No-Gaze RL.

\begin{figure}[t]
    \centering
    \includegraphics[width=\linewidth]{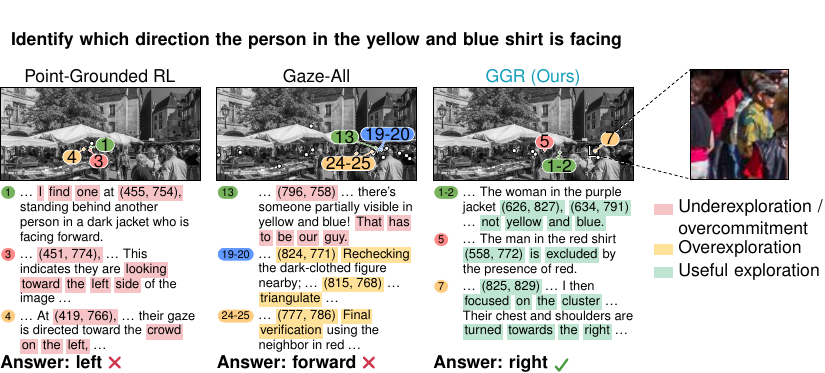}
    \vspace{-1em}
    \caption{\textbf{Example of \model{} eliciting useful grounded exploration compared to baselines.} On this example, Point-Grounded RL grounds few locations and commits to a location, Gaze-All re-evaluates many locations, and \model{} explores relevant locations before finding the correct answer.}
    \label{fig:behavioral-example}
\end{figure}

On the other hand, gaze guidance reduces result verification and re-examination compared to the condition without gaze guidance. Result verification and re-examination change by $-6.6$ and $+2.8$ percentage points, respectively, for \modelrl{} compared with $+25.7$ and $+15.7$ for No-Gaze RL (Figure~\ref{fig:behavioral-contrasts}a; $p<0.001$). These findings suggest that \modelrl{} does more visual exploration before arriving at an answer, while the No-Gaze-RL condition spends more time verifying an already established answer. We provide an example of this behavior difference in Figure~\ref{fig:behavioral-example}.

\begin{findingbox}
\modelrl{} does more visual exploration before arriving at an answer, while the No-Gaze-RL condition arrives at an answer quickly but spends more reasoning effort verifying the answer.
\end{findingbox}

\section{Discussion}
\label{sec:conclusion}

In this work, we investigated how human gaze can be transferred to modern vision-language models. Across chart understanding, counting, and
visual search, we find that gaze is most useful as a spatial prior for grounded reasoning. Our findings motivate \model{}, which uses a strong model to annotate reasoning conditioned on the gaze locations. \model{} obtains strong visual reasoning task performance while being more aligned with human gaze and producing more reliable visual claims compared to grounded reasoning baselines that do not use gaze guidance. 

Our results suggest that human behavior can be useful to a system with very different perceptual and cognitive constraints, but how the behavior is transferred matters. Supervision that annotated every fixation with a reasoning step (Gaze-All) significantly underperformed using a subset of gaze (Gaze-Subset) and using gaze as a prior (GGR). This suggests that the effectiveness of human guidance depends on how it is transferred and represented during training (Section~\ref{sec:gaze-supervision-strategies}). A sequence of fixations is a useful learning signal, but requiring a reasoning step for each fixation in a grounded reasoning trace is an ineffective way to transfer that signal to modern VLMs. This aligns with prior work that uses an abstraction of human behavior, rather than the raw behavior itself, as the supervision signal~\citep{saran2021efficientlyguidingimitationlearning,thammineni2023selective,salamatian2025chartgazeenhancingchartunderstanding}. Humans have attention and memory constraints that are missing from modern VLMs, and these constraints may require selective use of the human behavior when transferring it to a VLM with different constraints. 

Our results also show that human guidance can leave persistent behavioral and grounding changes after further training without gaze. After doing RL on GGR-SFT, we find persistent grounding accuracy and behavior changes compared to running the same method but without gaze guidance (Sections~\ref{sec:ggr-baselines} and~\ref{sec:behavior}). This is consistent with prior work showing that reinforcement learning amplifies and shapes behaviors already present in the base model~\citep{gandhi2025,zhao2504echo}. We show that gaze-guided supervised learning improves grounding reliability and amplifies visual exploration, and these behaviors persist after subsequent reinforcement learning.

To understand grounded reasoners in greater depth, we argue for measuring multiple metrics beyond accuracy, as similar accuracy does not establish similar strategies. We investigate multiple dimensions of model output in our work, including task success, alignment to gaze behavior, grounded reasoning reliability, and chain of thought reasoning behavior. We show that optimizing for one of these metrics does not imply predictable changes in the other metrics. For example, while finetuning on the raw gaze trace improved gaze alignment significantly, task performance dropped below the base model it was trained on. In fact, we show that improved reasoning trace quality (Section~\ref{sec:ggr-baselines}) is not reflected in accuracy compared to the baseline without gaze, consistent with prior work showing dissociations between task performance and model behavior~\citep{geirhos2020beyond,fel2022harmonizing}. Task accuracy alone offers a narrow perspective on visual reasoning behavior.

Several areas for future work remain. While we examined multiple base models (Section~\ref{sec:gaze-supervision-strategies}), we did not fully investigate how different pretraining data or architectures interact with the usefulness of the gaze supervision. Furthermore, the VLMs tested have significantly different constraints relative to human observers. Future work should examine how biologically-plausible attention and working memory constraints change the effectiveness of gaze supervision. 

\clearpage
\textbf{Acknowledgments.}
This work was supported by Princeton Research Computing and Princeton Language and Intelligence (PLI) resources, including the Della high-performance computing cluster.

\textbf{Ethics statement.}
This work involves research with human participants. The eye-tracking study
was conducted under a protocol approved by an institutional review board
(IRB). All participants provided informed consent before participation and
were informed that participation was voluntary and that they could withdraw
at any time.

\bibliographystyle{plainnat}
\bibliography{ref}

\setlength{\textfloatsep}{8pt plus 2pt minus 2pt}
\clearpage
\appendix
\renewcommand{\floatpagefraction}{0.95}
\section{Appendix}

\begingroup
\hypersetup{linkcolor=black}
\etocstandardlines
\etocsettocstyle{}{}
\etocsetnexttocdepth{subsection}
\localtableofcontents
\endgroup

\subsection{Additional Human Data Collection Details}\label{sec:app_human_data_coll}

\begin{figure}[!htbp]
    \centering
    \includegraphics[width=0.8\linewidth]{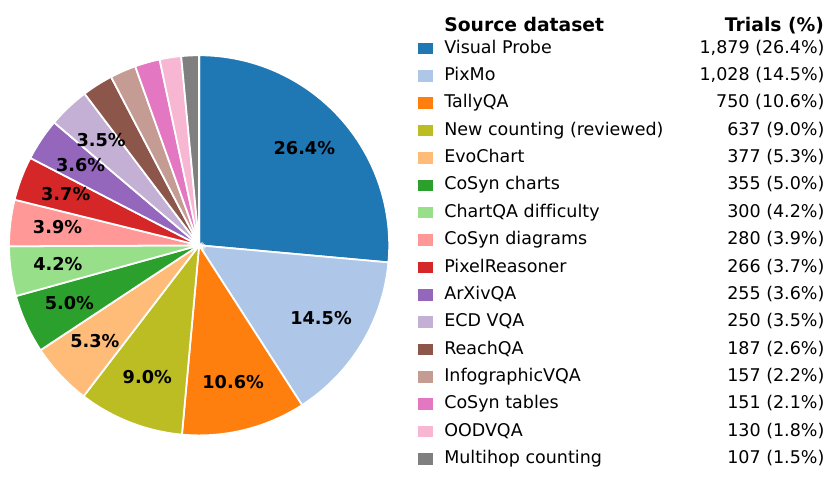}
    \caption{\textbf{Source-dataset composition} ($N=7,109$; 6,411 training and 698 validation trials). Counts and percentage shares are shown for each source.}
    \label{fig:dataset-breakdown}
\end{figure}

\textbf{Source datasets.} Figure~\ref{fig:dataset-breakdown} shows the source-dataset composition of 7,109 trials, with 6,411 used for training and 698 for validation. We had 58 participants in total and compensated \$20 per hour.

\textbf{Data collection.} English-speaking participants with normal or corrected-to-normal vision were recruited under an institutional review board-approved protocol. All participants provided informed consent and were informed that participation was voluntary and that they could withdraw at any time. Stimuli were presented using PsychoPy on an Asus VG27AQ LCD monitor operating at $2560 \times 1440$ pixels and 120~Hz. Participants viewed the display from approximately 89.5~cm with their head stabilized using a chin rest. Right-eye gaze was recorded using an SR Research EyeLink 1000 Plus desktop system at 1,000~Hz. Calibration and validation were performed before each session using the standard nine-point EyeLink array, with drift-correction trials interleaved every 10--20 task trials.

Participants were instructed to verbalize their reasoning and final answer before proceeding to the next trial. Trial audio was recorded and transcribed using Whisper large-v3~\citep{whisper}, with word-level timestamps. These verbalizations were not used to generate training traces because participants varied substantially in how extensively they explained their reasoning.

Eye-tracking recordings were processed at the fixation-event level. Finalized EDF files were converted to an ASCII event representation using EyeLinkIO (version 0.3.0) and the EyeLink EDF API, retaining the fixation onset times, durations, and mean screen-space gaze positions recorded by EyeLink. No additional offline fixation detection, duration filtering, or fixation merging was applied. Recorded synchronization markers aligned the PsychoPy and tracker clocks, and fixation times were expressed relative to trial onset. Fixations ending before trial onset were discarded, while those spanning trial onset were truncated to their within-trial duration.

To place gaze from different viewing states in a common reference frame, screen-space fixation positions were mapped to original-image coordinates by inverting the stimulus display geometry and applying the logged zoom level and viewport center active at fixation onset. Fixations mapped outside the original image were discarded, and retained coordinates were rounded to image pixels. No additional interpolation of missing gaze samples or explicit blink-window filtering was performed. Trials explicitly flagged as bad were excluded; participant-level exclusions were made only when manual inspection indicated particularly poor eye-tracking data quality.

\subsection{Model Annotation Details}\label{sec:app_annotation}

\subsubsection{GGR: Single-Stage Annotation}
For \model{}, the teacher receives the image $I$, question $q$, and fixation coordinate list $F$, and directly generates a grounded reasoning trace in a single stage. The fixation list serves as a spatial prior where the teacher may omit supplied fixations and ground additional locations when needed. The reference answer $a^\ast$ is withheld during trace generation.

Listing~\ref{sec:ggr-annotation-prompt} gives the gaze-conditioned prompt.

\begin{figure}[!htbp]
\tcbset{promptstylefmt/.append style={breakable=false}}
\begin{promptbox}{GGR: Gaze-Conditioned Annotation Prompt}{sec:ggr-annotation-prompt}
You are given:
\begin{promptenumerate}
\item an image,
\item a question,
\item human fixation points recorded while a person solved the task.
\end{promptenumerate}

Your goal is to generate a visually grounded reasoning trace with inline grounding coordinates that supports the final answer.

\medskip
IMPORTANT: Treat the human fixation points only as a SOFT SPATIAL PRIOR over where useful visual evidence may be located.

\medskip
The fixation points are NOT target grounding coordinates and are NOT in one-to-one correspondence with reasoning steps or objects.

\medskip
Use the fixation points to guide where you inspect the image, but determine every grounding coordinate from the IMAGE ITSELF.

\medskip
For each reasoning step:
\begin{promptenumerate}
\item Identify the actual visual entity or region that supports the claim.
\item Place the grounding coordinate on that entity or region.
\item Do not copy or reuse a fixation coordinate unless it is also an appropriate visual grounding.
\item If the fixation information and the visible image disagree, trust the image.
\item Do not force every fixation to appear in the reasoning.
\item Do not make up grounding coordinates that are not supported by the visible image just to support the final answer.
\end{promptenumerate}

\medskip
Output:
\begin{promptitemize}
\item Output only the visible reasoning trace with inline grounding coordinates formatted as (x, y) followed by exactly one \pboxed{<answer>}.
\item Do not make any reference to the 'fixation list' in your final output.
\end{promptitemize}
\end{promptbox}
\end{figure}

\subsubsection{No-Gaze SFT: Single-Stage Annotation}
For No-Gaze SFT, the teacher receives the image $I$ and question $q$, and directly generates a grounded reasoning trace in a single stage without human fixation guidance.

Listing~\ref{sec:no-gaze-annotation-prompt} gives the no-fixation prompt.

\begin{figure}[!htbp]
\tcbset{promptstylefmt/.append style={breakable=false}}
\begin{promptbox}{No-Gaze SFT: Annotation Prompt}{sec:no-gaze-annotation-prompt}
You are given:
\begin{promptenumerate}
\item an image, and
\item a question.
\end{promptenumerate}

Your goal is to answer the question from the visible image with a visually grounded reasoning trace and inline grounding coordinates.

\medskip
For each reasoning step:
\begin{promptenumerate}
\item Identify the actual visual entity or region that supports the claim.
\item Place the grounding coordinate on that entity or region.
\item Trust the visible image when determining the evidence and coordinates.
\item Do not make up visual evidence or grounding coordinates.
\end{promptenumerate}

\medskip
Output:
\begin{promptitemize}
\item Output only the visible reasoning trace with inline grounding coordinates formatted as (x, y) followed by exactly one \pboxed{<answer>}.
\end{promptitemize}
\end{promptbox}
\end{figure}

\subsubsection{Gaze-All: Three-Stage Annotation}
Gaze-All uses three annotation stages. First, the annotation model assigns a visual description to each candidate fixation in $F$, identifying the object or region at that location. Second, conditioned on the question $q$, it annotates the task relevance of these regions, such as identifying a target, ruling out a distractor, or checking for an omitted object. Third, the model uses the image, question, and annotated regions to generate candidate reasoning traces, with an accompanying reasoning step for every fixation. We exclude the participant's raw think-aloud transcript from this generation stage and withhold the reference answer $a^\ast$, requiring the model to derive its answer from the visual evidence.

\subsubsection{Gaze-Subset: Three-Stage Annotation}
Gaze-Subset uses the same three-stage annotation procedure as Gaze-All, but the final stage generates a reasoning trace using a strict subset of the supplied fixations, with no additional grounding points. Only Gaze-All and Gaze-Subset use the three-stage procedure; \model{} and No-Gaze SFT generate traces directly from their respective inputs.

Listing~\ref{sec:visual-reasoning-trace-prompt} gives the final-stage trace-generation prompt for Gaze-Subset.

\begin{figure}[!htbp]
\tcbset{promptstylefmt/.append style={breakable=false}}
\begin{promptbox}{Gaze-Subset: Final-Stage Reasoning Trace Prompt}{sec:visual-reasoning-trace-prompt}
Write a task-directed visual reasoning trace from the image, question, and annotated human fixation points.

\promptrule
\pbold{POINT SELECTION}
\begin{promptitemize}
\item Select useful supplied points. Omit others and use them in any semantic order.
\item Cite at least one supplied point using its \pcode{<N>} placeholder.
\item Cite individual points, not coordinate ranges or grouped IDs.
\item Describe the scene directly; do not narrate the human scanpath or annotation process.
\end{promptitemize}

\promptrule
\pbold{REASONING}
\begin{promptitemize}
\item Ground observations in the image and keep them relevant.
\item Verify object identity, attributes, counts, and relationships.
\item A correction or one verification pass is acceptable. Avoid repeated recounting, irrelevant exploration, and loops.
\item Never mention reference answers, judges, or previous attempts.
\end{promptitemize}

\promptrule
\pbold{OUTPUT}

Return only the reasoning trace followed by exactly one \pboxed{answer}. Do not include \pcode{<think>} tags or text after the answer.

\medskip
Image: \pvar{image}\\
Question: \pvar{question}\\
Annotated fixations:\\
\pcode{<1>} \pvar{visual label} --- \pvar{task-related function}\\
\pcode{<2>} \pvar{visual label} --- \pvar{task-related function}\\
...
\end{promptbox}
\end{figure}

\subsubsection{Rejection and Visual Verification}
We apply the same filtering procedure to candidate traces from \model{}, No-Gaze SFT, Gaze-All, and Gaze-Subset, using answer checks and image-grounded assessments of reasoning quality. The checks remove candidates with incorrect answers, invalid point references, excessive repetition, or commentary about the annotation process. A subsequent visual assessment checks whether the trace identifies the correct objects and attributes, supports its claims with appropriate spatial references, and avoids counting errors, contradictions, or unsupported conclusions. Accepted traces are converted into training targets containing the grounded reasoning and final answer.

\vspace{-8pt}
\subsection{Baseline Details}
\vspace{-4pt}

\textbf{Raw-Gaze-SFT/RL.} We process the raw fixation sequence and think aloud into a single reasoning chain. We create a flattened list of (x, y) coordinates and interleave the think aloud in between the coordinate list at locations that align based on the Whisper transcriptions. Each flattened list + think aloud + final answer is converted into a training target.

\textbf{ViGoRL-MCTS.} Following~\citet{sarch2026groundedreinforcementlearningvisual}, we generated spatially grounded reasoning trajectories with Qwen3.5-122B-A10B using Monte Carlo Tree Search (MCTS), with 16 simulations per example, four children per expansion, and four rollouts per child. The teacher received the image and question and generated reasoning steps grounded in image coordinates, while reference answers were used to score completed rollouts and guide search. We retained answer-correct trajectories and constructed correction examples by pairing unsuccessful and successful continuations from the same node. After deduplication, we selected up to 25 trajectories per example.

\subsection{Training and Evaluation Details}\label{sec:app_train_eval_details}

\paragraph{SFT Training Protocol.}
All conditions use the same system prompt (Listing~\ref{sec:sft-rl-system-prompt}) during training and evaluation.
We fine-tune using supervised next-token prediction on the reasoning trace and final answer. Training uses full-parameter fine-tuning of the language model with the vision encoder frozen. 

\begin{table}[!htbp]
\centering
\small
\begin{tabular}{ll}
\hline
Setting & Configuration \\
\hline
Learning rate & $5\times10^{-6}$ \\
Schedule & Cosine decay with 3\% warmup \\
Weight decay & $0$ \\
Gradient clipping & Max norm $1.0$ \\
Training epochs & 5 \\
Batch size & 64 \\
Maximum sequence length & 32,768 tokens \\
Image pixel bounds & 921,600--8,388,608 \\
Precision / parallelism & BF16 / DeepSpeed ZeRO-3 across 8 GPUs \\
\hline
\end{tabular}
\caption{SFT training configuration.}
\label{tab:sft_setup}
\end{table}

\begin{figure}[!htbp]
\tcbset{promptstylefmt/.append style={breakable=false}}
\begin{promptbox}{SFT and RL System Prompt}{sec:sft-rl-system-prompt}
You are a helpful, conversational assistant tasked with answering a question about an image. Generate a reasoning trace with reasoning grounded in points in the image ((x, y) coordinates) to help you answer.

\begin{promptenumerate}
\item \pbold{Reasoning}: A detailed, free-flowing chain of thought enclosed in \pcode{<think>} and \pcode{</think>} tags, with reasoning grounded in points in the image ((x, y) coordinates).

While reasoning, explicitly model these cognitive behaviors:
\begin{promptitemize}
\item \pbold{Planning}: Outline possible approaches before committing.
\item \pbold{Exploration}: Consider multiple image regions or interpretations.
\item \pbold{Evaluation}: Compare alternatives and verify against visual evidence.
\item \pbold{Reflection}: Revisit earlier ideas if they might still be viable.
\item \pbold{Error Correction}: Acknowledge mistakes if grounded reasoning reveals an incorrect path.
\end{promptitemize}
\item \pbold{Response}: After reasoning and reaching a conclusion, generate a clear, conversational response. After the response, indicate your final answer using \pboxed{}.
\end{promptenumerate}
\end{promptbox}
\end{figure}

\paragraph{RL Training Protocol.}
We use the system prompt in Listing~\ref{sec:sft-rl-system-prompt}.
We train with GSPO for 400 training steps. Each batch contains 256 prompts balanced across chart, counting, and search tasks. Advantages do not use standard-deviation normalization. We do not apply KL regularization nor an entropy bonus. We use full-finetuning of all parameters. RL hyperparameters are detailed in Table~\ref{tab:rl_setup}.

The reward combines task-specific answer correctness (weight $0.95$) and valid point-grounded reasoning format (weight $0.05$); invalid formatting also zeros the correctness component. For counting, the correctness reward is $2^{-|\hat{a}-a^\ast|}$ for absolute errors below four and zero otherwise. 

\begin{table}[!htbp]
\centering
\small
\begin{tabular}{ll}
\hline
Setting & Configuration \\
\hline
Learning rate & 20-step warmup to $2\times10^{-5}$; \\
              & linear decay to $7\times10^{-6}$ at step 400 \\
Prompt batch / optimization minibatch & 256 / 128 \\
Responses per prompt & 8 \\
GSPO clipping & $\epsilon_{\mathrm{low}}=3\times10^{-4}$, $\epsilon_{\mathrm{high}}=4\times10^{-4}$ \\
Rollout sampling & Temperature $1.0$, top-$p$ $1.0$ \\
Maximum prompt / response length & 16,384 / 14,336 tokens \\
Precision / parallelism & BF16 / FSDP2 across 8 GPUs \\
\hline
\end{tabular}
\caption{RL training configuration.}
\label{tab:rl_setup}
\end{table}

\subsection{Evaluation Protocol.}

We detail our evaluation protocol in Table~\ref{tab:eval_setup}. We use lmms-eval~\citep{zhang2025lmms} for all evaluations.

\begin{table}[!htbp]
\centering
\small
\begin{tabular}{lp{0.68\linewidth}}
\hline
Setting & Configuration \\
\hline
Chart benchmarks & ChartQA, ChartQA Pro \\
Counting benchmarks & CountQA, CountBenchQA \\
Search benchmarks & MME-RealWorld-Lite (Vero subsample), Visual Probe \\
Decoding & Temperature $1.0$, top-$p$ $0.95$, top-$k$ $20$, presence penalty $1.5$ \\
Output budget & 16,384 tokens \\
Image preprocessing & Min 2M and max 10M \\
\hline
\end{tabular}
\caption{Evaluation configuration.}
\label{tab:eval_setup}
\end{table}

\paragraph{Behavioral Analysis.}
We analyze reasoning traces from held-out gaze evaluations using Qwen3.5-122B-A10B as a text-only behavioral coder, with temperature zero and thinking disabled. The coder counts distinct, explicitly expressed occurrences in each rollout. We report two category scores and four individual behaviors (Table~\ref{tab:behavior_categories}). We average three rollouts per example and report paired differences from the base model (mean $\pm$ SEM across $698$ examples).

\begin{figure}[!htbp]
    \centering
    \includegraphics[width=0.6\linewidth,trim=0bp 0bp 313bp 0bp,clip]{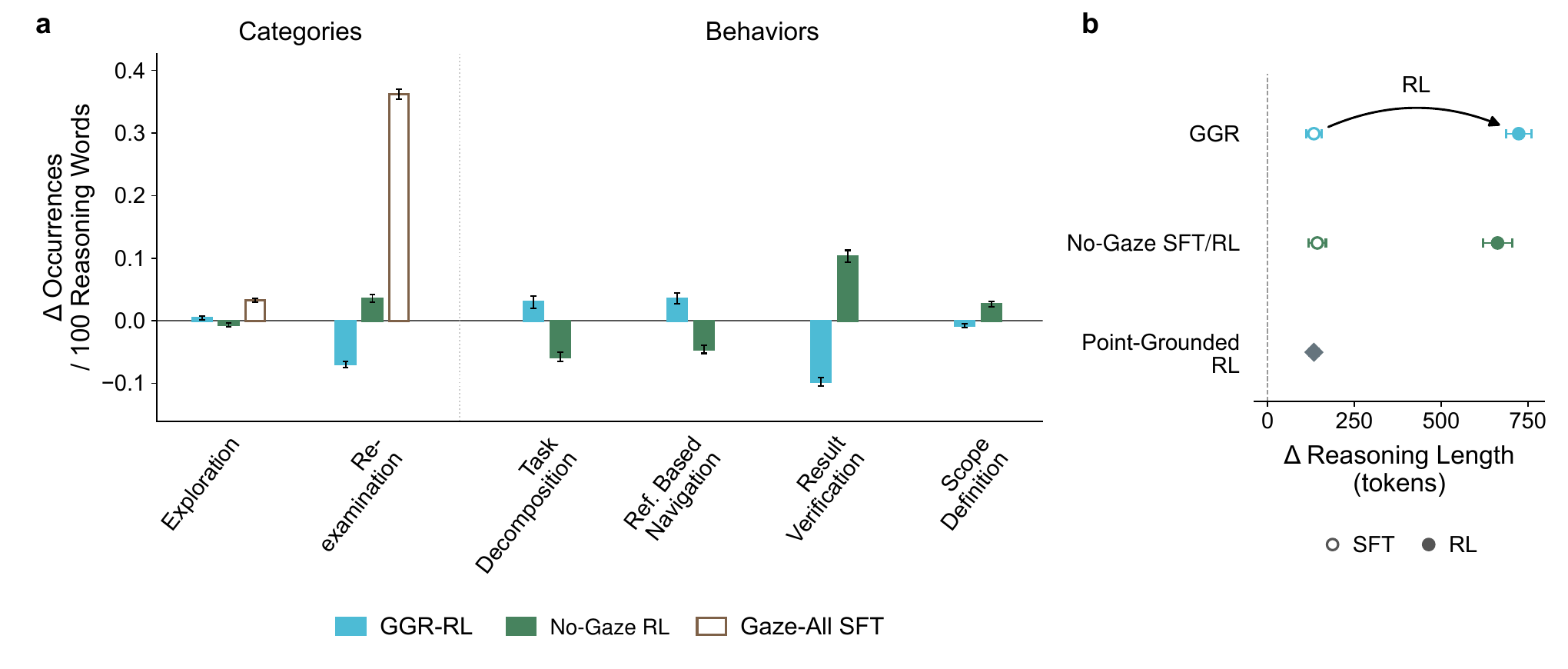}
    \caption{Behavioral contrasts relative to base in occurrences per 100 reasoning words. Categories, individual behaviors, and conditions match Figure~\ref{fig:behavioral-contrasts}a, which reports behavior presence. Reported is mean $\pm$ SEM across paired example differences after averaging three rollouts per example ($N=698$).}
    \label{fig:behavioral-full-appendix}
\end{figure}

\begin{table}[!htbp]
\centering
\small
\begin{tabular}{p{0.28\linewidth}p{0.66\linewidth}}
\hline
Category / behavior & Coded behaviors / definition \\
\hline
Exploration & Exploring a different image region for visual evidence. \\
Re-examination & Revisiting of earlier observations and verification of provisional or final results. \\
Task decomposition & Dividing the task into organized intermediate steps or milestones. \\
Reference-based navigation & Uses a landmark or salient object to orient the search. \\
Result verification & Checks a provisional or final result for completeness, consistency, arithmetic, and compliance with the task. \\
Scope definition & Defines inclusion, exclusion, boundary, or ambiguity rules before processing. \\
\hline
\end{tabular}
\caption{Plotted behavioral categories and individual behaviors. Counts describe behaviors expressed in generated reasoning, rather than independently verified visual actions.}
\label{tab:behavior_categories}
\end{table}

\vspace{-8pt}
\subsection{Fixation Alignment Metric.}\label{sec:app_fix_align}
For example $i$, let $h_{ij}$ denote human fixation $j$, and let
$p_{irkt}$ denote grounding point $k$ from model response $r$ at
checkpoint $t$. Coordinates lie in the normalized image plane
$[0,1000]^2$. After excluding invalid and out-of-bounds coordinates,
we construct
\[
q_{cirt}(x)=
(1-\epsilon_c)\frac{1}{M_{irt}}
\sum_{k=1}^{M_{irt}}
\mathcal{N}(x;p_{irkt},\sigma_c^2 I)
+\epsilon_c u,
\qquad u=10^{-6},
\]
where $c$ indexes training condition and $M_{irt}$ is the number
of valid reported points. 

We score each response by its mean log-likelihood relative to uniform:
\[
A_{cirt}=
\frac{1}{N_i}\sum_{j=1}^{N_i}
\log_2\frac{q_{cirt}(h_{ij})}{u}.
\]

We jointly fit $(\sigma_c,\epsilon_c)$ separately for each
condition using 336 calibration examples.
The candidate grid is
\[
\sigma\in\{25,35,50,70,100,140,200\},\qquad
\epsilon\in\{0.01,0.03,0.1,0.2,0.4,0.7,0.9\}.
\]
Selection maximizes calibration-set mean alignment.

\paragraph{Time-resolved and order-aware alignment.}
Our spatial alignment score does not take the ordering of fixations into account. The below three analyses extend this score to include fixation order. 
 
\emph{Time-resolved alignment.} We split the sequence of fixations into five ordered parts with equal numbers of fixations, and compute the alignment score on each part. Here, we discard the first two fixations (which reflect the central starting position) and exclude scanpaths with fewer than ten remaining fixations. 

\emph{Order score.} We compute the Goodman--Kruskal $\gamma$ between fixation time and the
position of the fixated region in the model's response ($1$ if regions are fixated in the
order the model mentions them, $-1$ for the reverse order). Here, a region is a grounding
point more than 50 units from every earlier region, each fixation is assigned to its nearest
region within 100 units, and we require at least three assigned fixations on at least two
regions.

\emph{ScanMatch-style similarity.} Following ScanMatch~\citep{cristino2010scanmatch}, we
align the sequence of regions with the sequence of human dwells (runs of consecutive
fixations within 50 units of their centroid) using the Needleman--Wunsch algorithm. Here, a
region and a dwell at distance $d$ score $1-d/150$. To separate order from spatial overlap, we repeat the
alignment with the regions shuffled (mean of ten shuffles) and report the difference per
region for responses with at least two regions.
 
\emph{Human reference.} For scanpaths with at least 24 fixations, we summarize the odd-numbered fixations by their ten largest dwells, ordered by first visit, and score this summary against the even-numbered fixations.

\FloatBarrier
\begingroup
\raggedbottom
\setlength{\intextsep}{3pt plus 1pt minus 1pt}
\setlength{\parskip}{4pt}
\subsection{Additional Experiments}\label{sec:app_add_results}

We compare models using paired permutation tests over source images. All p-values are two-sided and Holm-corrected for multiple comparisons.

\paragraph{Condition-fitted gaze alignment.}
Figure~\ref{fig:condition-fitted-ig} reports alignment across supervision conditions before and after RL, complementing the precision and coverage analysis in Figure~\ref{fig:gaze-alignment-overview}a.

\begin{figure}[H]
    \centering
    \includegraphics[width=0.65\linewidth]{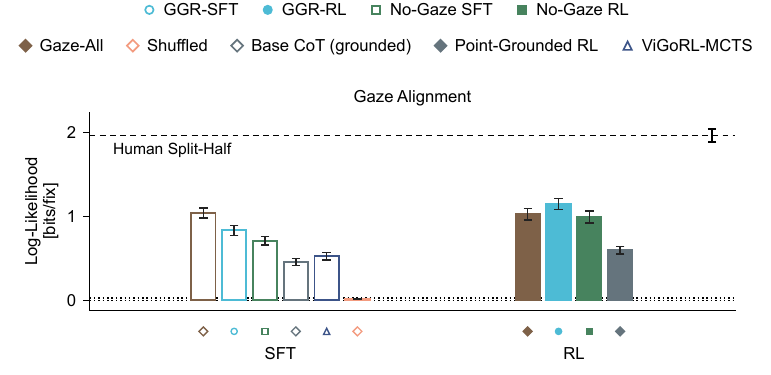}
    \caption{\textbf{Condition-fitted gaze alignment.} Alignment after SFT (open bars) and after RL (filled bars). The dashed line denotes human split-half alignment; the dotted line denotes the uniform reference. Reported is mean $\pm$ SEM.}
    \label{fig:condition-fitted-ig}
\end{figure}

\paragraph{Time-resolved and order-aware gaze alignment.}
With shared parameters (Appendix~\ref{sec:app_fix_align}), \model{} is more aligned than
No-Gaze overall (Figure~\ref{fig:alignment-time}a) and in every fifth of the human scanpath
(Figure~\ref{fig:alignment-time}b--c; all $p<0.001$).
ScanMatch-style similarity is also higher for \model{} (SFT and RL both $p<0.001$).

We additionally compare gaze correspondence in teacher-generated reasoning supervision on 100 trials held out from training. Using the same teacher model, we compare GGR with Gazette~\citep{mondal2026gaze} to obtain reasoning annotations. GGR achieves higher ScanMatch similarity (0.172 versus 0.103; paired difference 0.069, 95\% CI [0.053, 0.085]) and lower human-to-trace spatial coverage distance (0.147 versus 0.207).

\begin{figure}[H]
    \centering
    \includegraphics[width=\linewidth]{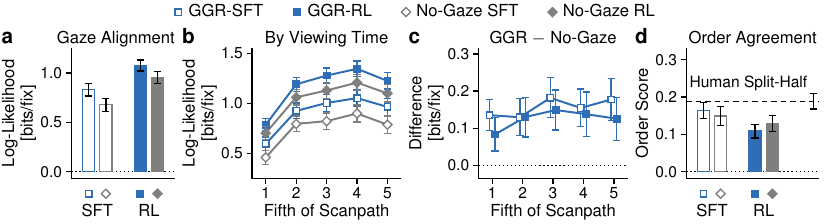}
    \caption{\textbf{Gaze alignment over viewing time and order agreement.} (a) Alignment with
shared parameters ($\sigma=70$, $\epsilon=0.4$). (b) Alignment within each fifth of the
human scanpath. (c) Paired \model{} minus No-Gaze difference in each fifth, with
bootstrapped 95\% CIs. (d) Order score. The dashed line denotes the human split-half
reference ($0.19$). Open and filled markers denote SFT and RL, respectively. Panels a, b,
and d report mean $\pm$ SEM.}
    \label{fig:alignment-time}
\end{figure}

Figure~\ref{fig:spatial-behavior} provides additional plots of \model{} and Point-Grounded RL across RL steps for unique points output, human fixation coverage, and human fixation precision on the held out set. Figure~\ref{fig:behavioral-full-appendix} reports the behavioral contrasts in Figure~\ref{fig:behavioral-contrasts}a in occurrences per 100 reasoning words.

\begin{figure}[H]
    \centering
    \includegraphics[width=\linewidth]{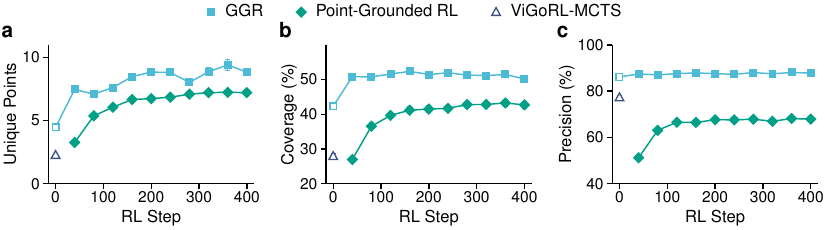}
    \caption{Spatial behavior across RL training. Reported is mean $\pm$ SEM across images. Open markers denote SFT, while filled markers denote RL.}
    \label{fig:spatial-behavior}
\end{figure}

\ifdefined\includevisualgroundingaudit
\paragraph{Visual grounding.}
Figure~\ref{fig:visual-grounding-sft} reports SFT differences; Table~\ref{tab:visual-grounding-rates} reports condition rates. We used GPT-6.1-Sol to assess one sampled point, when present, and its associated claim. We provided the question, image, marked crop, and surrounding text. We assessed placement as on, near, off, or unclear, and the claim as supported, contradicted, or unclear. For joint grounding, we required a supported claim and a point on or near the target.

Two participants judged the same 100 outputs from 25 questions across six benchmarks. They could not see the conditions, reference answers, or assistant ratings. We compare their judgments to GPT-6.1-Sol in Table~\ref{tab:human-grounding-agreement}.

\begin{table}[H]
    \centering
    \begingroup
    \small
    \setlength{\tabcolsep}{5pt}
    \renewcommand{\arraystretch}{1.15}
    \begin{tabular}{@{}lccc@{}}
        \toprule
        Measure & H1 vs.\ assistant & H2 vs.\ assistant & H1 vs.\ H2 \\
        \midrule
        Point on/near target & 90 (0.534) & 90 (0.561) & 92 (0.512) \\
        Claim: supported, contradicted, unclear & 89 (0.437) & 87 (0.306) & 91 (0.275) \\
        Supported claim + point on/near target & 86 (0.572) & 84 (0.525) & 90 (0.608) \\
        \bottomrule
    \end{tabular}
    \endgroup
    \caption{\textbf{Agreement with humans.} Agreement (\%) with Cohen's $\kappa$ in parentheses.}
    \label{tab:human-grounding-agreement}
\end{table}

\par\addvspace{\intextsep}\noindent
\begin{minipage}{\linewidth}
    \captionsetup{hypcap=false}
    \centering
    \setlength{\unitlength}{\dimexpr\linewidth/1250\relax}
    \begin{minipage}[t]{0.65\linewidth}
    \vspace{0pt}
    \begingroup
    \fontsize{23\unitlength}{28\unitlength}\fontfamily{phv}\selectfont
    \colorlet{vgheadingbg}{resultshighlight} 
    \newcommand{\vgheading}[1]{%
        \makebox[0pt][l]{\smash{\color{vgheadingbg}\rule[-7\unitlength]{\linewidth}{30\unitlength}}}%
        #1%
    }
    \begin{minipage}[t]{0.6153846\linewidth}
        \vspace{0pt}
        \setlength{\parskip}{0pt}
        \noindent\strut Condition rates (\%, 95\% CI)\par
        \setlength{\tabcolsep}{3\unitlength}
        \setlength{\abovetopsep}{0pt}
        \setlength{\aboverulesep}{2\unitlength}
        \setlength{\belowrulesep}{3\unitlength}
        \setlength{\heavyrulewidth}{1.2\unitlength}
        \setlength{\lightrulewidth}{1\unitlength}
        \renewcommand{\arraystretch}{1}
        \begin{tabular*}{\linewidth}{@{\extracolsep{\fill}}lrr@{}}
            \toprule
            \multicolumn{1}{@{}c}{Stage} & \multicolumn{1}{c}{No-Gaze} & \multicolumn{1}{c@{}}{GGR} \\[8\unitlength]
            \midrule
            \multicolumn{3}{@{}l@{}}{\vgheading{Supported claim + point on/near target}} \\[6\unitlength]
            SFT     & 58.7 [56.5, 60.9] & \textbf{63.3 [61.0, 65.5]} \\[6\unitlength]
            Post-RL & 70.5 [68.4, 72.6] & \textbf{74.9 [72.8, 77.0]} \\
            \addlinespace[3\unitlength]
            \multicolumn{3}{@{}l@{}}{\vgheading{Point on/near target}} \\[6\unitlength]
            SFT     & 68.3 [66.1, 70.3] & \textbf{72.7 [70.7, 74.8]} \\[6\unitlength]
            Post-RL & 79.8 [77.9, 81.7] & \textbf{84.1 [82.3, 85.8]} \\[4\unitlength]
            \bottomrule
        \end{tabular*}
    \end{minipage}
    \endgroup
    \captionof{table}{\textbf{Visual-grounding condition rates.} Judgements of 1,500 matched questions (250 per benchmark; 6,000 outputs). The evaluation assesses one randomly sampled coordinate occurrence and its local claim. We report average (\%) and 95\% CI. Bold marks the higher rate in each row.}
    \label{tab:visual-grounding-rates}
    \end{minipage}\hfill
    \begin{minipage}[t]{0.27\linewidth}
        \vspace{0pt}
        \centering
        \includegraphics[width=\linewidth,trim=0bp 0bp 0bp 30bp,clip]{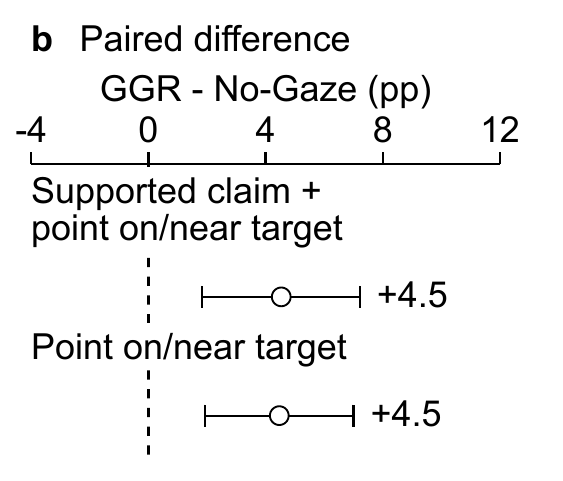}
        \captionof{figure}{\textbf{GGR improves visual grounding quality compared to the no gaze condition.} Paired GGR minus No-Gaze differences in visual grounding after SFT. Reported is mean $\pm$ bootstrapped 95\% CI.}
        \label{fig:visual-grounding-sft}
    \end{minipage}
\end{minipage}
\par\addvspace{\intextsep}

\fi

\newlength{\appendixtablewidth}
\setlength{\appendixtablewidth}{\textwidth}
\ifdim\appendixtablewidth>5.4in
    \setlength{\appendixtablewidth}{5.4in}
\fi
\paragraph{Benchmark-level results for gaze supervision.}
Table~\ref{tab:sft-rl-task-results} expands Table~\ref{tab:gaze-supervision-strategies} with results on each of the six benchmarks after SFT and RL.

\begin{table}[H]
    \centering
    \begingroup
    \setlength{\tabcolsep}{3pt}
    \renewcommand{\arraystretch}{1.10}
    \resizebox{\appendixtablewidth}{!}{    \begin{tabular}{@{}l*{8}{c}@{}}
    \hline
    & \multicolumn{2}{c}{\textbf{Chart}} & \multicolumn{2}{c}{\textbf{Counting}} & \multicolumn{2}{c}{\textbf{Search}} & & \\
    \cmidrule(lr{3pt}){2-3}\cmidrule(lr{3pt}){4-5}\cmidrule(lr{3pt}){6-7}
    \noalign{\vskip 3pt}
    \textbf{Condition} & \shortstack{ChartQA\\Pro} & ChartQA & \shortstack{Count\\BenchQA} & CountQA & \shortstack{MME\\RealWorld\\Lite} & \shortstack{Visual\\Probe} & \textbf{Overall} & \shortstack{\textbf{Gaze}\\\textbf{Alignment}\\{[bits/fix]}} \\
    \hline
    \noalign{\vskip 4pt}
    \rowcolor{resultshighlight}[0pt][0pt]
    \multicolumn{9}{@{}l@{}}{\rule[-3pt]{0pt}{16pt}\textbf{Supervised fine-tuning (SFT)}} \\
    \noalign{\vskip 2pt}
    \multicolumn{9}{@{}l}{\textit{Controls}} \\
    No-Gaze & $\mathbf{59.6}^{\pm 0.1}$ & $\mathbf{90.5}^{\pm 0.0}$ & $\mathbf{94.9}^{\pm 0.5}$ & $40.5^{\pm 0.3}$ & $54.2^{\pm 0.2}$ & $45.6^{\pm 0.9}$ & $64.2^{\pm 0.2}$ & $0.71^{\pm 0.05}$ \\
    Shuffled & $49.1^{\pm 0.3}$ & $89.2^{\pm 0.1}$ & $82.6^{\pm 2.0}$ & $20.2^{\pm 0.5}$ & $51.5^{\pm 0.2}$ & $42.5^{\pm 1.1}$ & $55.8^{\pm 0.4}$ & $0.01^{\pm 0.00}$ \\
    \hline
    \multicolumn{9}{@{}l}{\textit{Human-gaze supervision}} \\
    Raw-Gaze & $31.2^{\pm 0.2}$ & $76.1^{\pm 0.2}$ & $85.0^{\pm 1.2}$ & $24.9^{\pm 0.3}$ & $41.9^{\pm 0.7}$ & $41.9^{\pm 0.7}$ & $50.2^{\pm 0.3}$ & $0.96^{\pm 0.06}$ \\
    Gaze-All & $50.2^{\pm 0.2}$ & $88.4^{\pm 0.2}$ & $83.3^{\pm 0.5}$ & $22.8^{\pm 0.1}$ & $50.8^{\pm 0.6}$ & $39.8^{\pm 0.8}$ & $55.9^{\pm 0.2}$ & $\mathbf{1.04}^{\pm 0.06}$ \\
    Gaze-Subset & $58.6^{\pm 0.1}$ & $\mathbf{91.0}^{\pm 0.2}$ & $93.3^{\pm 0.4}$ & $32.9^{\pm 0.4}$ & $55.2^{\pm 0.8}$ & $46.0^{\pm 1.0}$ & $62.8^{\pm 0.2}$ & $0.82^{\pm 0.05}$ \\
    \modelsft{} & $58.4^{\pm 0.5}$ & $\mathbf{90.8}^{\pm 0.0}$ & $\mathbf{95.0}^{\pm 0.4}$ & $\mathbf{41.8}^{\pm 0.3}$ & $\mathbf{57.7}^{\pm 0.1}$ & $\mathbf{48.0}^{\pm 0.7}$ & $\mathbf{65.3}^{\pm 0.2}$ & $0.84^{\pm 0.06}$ \\
    \hline
    \noalign{\vskip 4pt}
    \rowcolor{resultshighlight}[0pt][0pt]
    \multicolumn{9}{@{}l@{}}{\rule[-3pt]{0pt}{16pt}\textbf{Reinforcement learning (RL)}} \\
    \noalign{\vskip 2pt}
    \multicolumn{9}{@{}l}{\textit{Controls}} \\
    No-Gaze-SFT $\rightarrow$ RL & $\mathbf{63.6}^{\pm 0.4}$ & $\mathbf{92.0}^{\pm 0.1}$ & $95.1^{\pm 0.0}$ & $\mathbf{52.9}^{\pm 0.1}$ & $57.3^{\pm 0.3}$ & $51.8^{\pm 0.2}$ & $\mathbf{68.8}^{\pm 0.1}$ & $0.99^{\pm 0.07}$ \\
    Shuffled & -- & -- & -- & -- & -- & -- & -- & -- \\
    \hline
    \multicolumn{9}{@{}l}{\textit{Human-gaze supervision}} \\

    Raw-Gaze & $59.8^{\pm 0.4}$ & $90.9^{\pm 0.3}$ & $89.7^{\pm 0.4}$ & $31.4^{\pm 0.5}$ & $54.4^{\pm 0.2}$ & $\mathbf{53.5}^{\pm 0.5}$ & $63.3^{\pm 0.1}$ & $1.02^{\pm 0.06}$ \\
    Gaze-All & $58.8^{\pm 0.2}$ & $\mathbf{91.9}^{\pm 0.1}$ & $93.3^{\pm 0.1}$ & $38.1^{\pm 0.5}$ & $58.1^{\pm 0.4}$ & $51.5^{\pm 0.5}$ & $65.3^{\pm 0.1}$ & $1.03^{\pm 0.07}$ \\
    Gaze-Subset & $61.1^{\pm 0.3}$ & $\mathbf{92.0}^{\pm 0.1}$ & $94.8^{\pm 0.1}$ & $45.0^{\pm 0.3}$ & $57.8^{\pm 0.1}$ & $\mathbf{53.1}^{\pm 0.7}$ & $67.3^{\pm 0.1}$ & $0.99^{\pm 0.07}$ \\
    \modelrl{} & $\mathbf{64.0}^{\pm 0.7}$ & $\mathbf{91.9}^{\pm 0.3}$ & $\mathbf{95.9}^{\pm 0.1}$ & $50.7^{\pm 0.5}$ & $\mathbf{59.6}^{\pm 0.2}$ & $50.8^{\pm 0.6}$ & $\mathbf{68.8}^{\pm 0.2}$ & $\mathbf{1.15}^{\pm 0.07}$ \\
    \hline
    \end{tabular}    }
    \endgroup
    \caption{\textbf{Benchmark-level comparison of gaze-supervision strategies.} Conditions, metrics, and bolding follow Table~\ref{tab:gaze-supervision-strategies}. Overall is the unweighted mean of the six benchmarks.}
    \label{tab:sft-rl-task-results}
    \label{tab:sft-task-results}
    \label{tab:rl-task-results}
\end{table}

\par\noindent
\begin{minipage}{\linewidth}
\paragraph{GGR ablations.}
Table~\ref{tab:ggr-component-ablations} compares \modelsft{} with ablated GGR supervision.

\begin{table}[H]
    \centering
    \begingroup
    \setlength{\tabcolsep}{6pt}
    \renewcommand{\arraystretch}{1.15}
    \resizebox{\appendixtablewidth}{!}{
    \begin{tabular}{@{}l*{6}{c}>{\columncolor{resultscolumnhighlight}}c>{\columncolor{resultscolumnhighlight}[\tabcolsep][0pt]}c@{}}
    \hline
    & \multicolumn{2}{c}{\textbf{Chart}} & \multicolumn{2}{c}{\textbf{Counting}} & \multicolumn{2}{c}{\textbf{Search}} & \multicolumn{2}{c@{}}{} \\
    \cmidrule(lr{3pt}){2-3}\cmidrule(lr{3pt}){4-5}\cmidrule(lr{3pt}){6-7}
    \noalign{\vskip 3pt}
    \textbf{Condition} & \shortstack{ChartQA\\Pro} & ChartQA & \shortstack{Count\\BenchQA} & CountQA & \shortstack{MME\\RealWorld\\Lite} & \shortstack{Visual\\Probe} & \textbf{Overall} & \shortstack{\textbf{Gaze}\\\textbf{Alignment}\\{[bits/fix]}} \\
    \hline
    \modelsft{} & $58.4^{\pm 0.5}$ & $90.8^{\pm 0.0}$ & $95.0^{\pm 0.4}$ & $41.8^{\pm 0.3}$ & $57.7^{\pm 0.1}$ & $48.0^{\pm 0.7}$ & $\mathbf{65.3}^{\pm 0.2}$ & $\mathbf{0.84}^{\pm 0.06}$ \\
    Coordinate-only & $34.79$ & $84.68$ & $93.28$ & $40.31$ & $48.83$ & $49.51$ & $58.57$ & $0.7784^{\pm 0.0542}$ \\
    Reasoning-only & $55.69$ & $89.92$ & $93.48$ & $32.46$ & $54.98$ & $44.70$ & $61.87$ & $0.0018^{\pm 0.0019}$ \\
    Shuffled & $56.9$ & $89.6$ & $95.1$ & $38.4$ & $55.1$ & $45.2$ & $63.4$ & $0.6799^{\pm 0.0513}$ \\
    \hline
    \end{tabular}
    }
    \endgroup
    \caption{\textbf{GGR-SFT ablations.} Task accuracy (\%) and human-gaze alignment. Overall is the unweighted mean of all six benchmarks.}
    \label{tab:ggr-component-ablations}
\end{table}
\end{minipage}
\par

\paragraph{SFT with additional base models.}
Tables~\ref{tab:gemma3-4b-sft-task-results} and~\ref{tab:mimo-sft-task-results} report SFT task performance and human-gaze alignment for Gemma 3 4B and MiMo-VL-7B under the conditions in Table~\ref{tab:gaze-supervision-strategies}.

\begin{table}[H]
    \centering
    \begingroup
    \setlength{\tabcolsep}{3pt}
    \renewcommand{\arraystretch}{1.15}
    \resizebox{\appendixtablewidth}{!}{%
    \begin{tabular}{@{}l*{8}{c}@{}}
    \hline
    & \multicolumn{2}{c}{\textbf{Chart}} & \multicolumn{2}{c}{\textbf{Counting}} & \multicolumn{2}{c}{\textbf{Search}} & & \\
    \cmidrule(lr{3pt}){2-3}\cmidrule(lr{3pt}){4-5}\cmidrule(lr{3pt}){6-7}
    \noalign{\vskip 3pt}
    \textbf{Condition} & \shortstack{ChartQA\\Pro} & ChartQA & \shortstack{Count\\BenchQA} & CountQA & \shortstack{MME\\RealWorld\\Lite} & \shortstack{Visual\\Probe} & \textbf{Overall} & \shortstack{\textbf{Gaze}\\\textbf{Alignment}\\{[bits/fix]}} \\
    \hline
    \multicolumn{9}{@{}l}{\textit{Base model}} \\
    Gemma 3 4B + CoT & $34.2^{\pm 0.5}$ & $70.1^{\pm 0.3}$ & $71.2^{\pm 0.1}$ & $13.3^{\pm 0.1}$ & $36.9^{\pm 0.3}$ & $20.0^{\pm 0.6}$ & $40.9^{\pm 0.2}$ & $-0.03^{\pm 0.00}$ \\
    \hline
    \multicolumn{9}{@{}l}{\textit{Human-fixation SFT}} \\
    Gaze-All & $26.4^{\pm 0.8}$ & $70.1^{\pm 0.2}$ & $60.4^{\pm 1.1}$ & $9.7^{\pm 0.5}$ & $34.9^{\pm 0.5}$ & $18.1^{\pm 0.8}$ & $36.6^{\pm 0.3}$ & $0.34^{\pm 0.03}$ \\
    Gaze-Subset & $30.1^{\pm 0.2}$ & $77.0^{\pm 0.2}$ & $73.9^{\pm 0.4}$ & $12.3^{\pm 0.1}$ & $35.3^{\pm 0.3}$ & $19.8^{\pm 1.0}$ & $41.4^{\pm 0.2}$ & $0.48^{\pm 0.04}$ \\
    \modelsft{} & $\mathbf{31.9}^{\pm 0.5}$ & $\mathbf{78.2}^{\pm 0.1}$ & $\mathbf{77.9}^{\pm 0.5}$ & $\mathbf{14.2}^{\pm 0.2}$ & $\mathbf{38.8}^{\pm 0.4}$ & $\mathbf{20.5}^{\pm 0.8}$ & $\mathbf{43.6}^{\pm 0.2}$ & $\mathbf{0.51}^{\pm 0.04}$ \\
    \hline
    \end{tabular}%
    }
    \endgroup
    \caption{\textbf{SFT task performance and gaze alignment with Gemma 3 4B.} Condition names follow Table~\ref{tab:gaze-supervision-strategies}. Task scores report Avg@3 ($\pm$ SEM). We bold SFT task means within 0.5 percentage points of the best SFT mean in each column, and the best available SFT gaze-alignment mean.}
    \label{tab:gemma3-4b-sft-task-results}
\end{table}

\begin{table}[H]
    \centering
    \begingroup
    \setlength{\tabcolsep}{3pt}
    \renewcommand{\arraystretch}{1.15}
    \resizebox{\appendixtablewidth}{!}{%
    \begin{tabular}{@{}l*{8}{c}@{}}
    \hline
    & \multicolumn{2}{c}{\textbf{Chart}} & \multicolumn{2}{c}{\textbf{Counting}} & \multicolumn{2}{c}{\textbf{Search}} & & \\
    \cmidrule(lr{3pt}){2-3}\cmidrule(lr{3pt}){4-5}\cmidrule(lr{3pt}){6-7}
    \noalign{\vskip 3pt}
    \textbf{Condition} & \shortstack{ChartQA\\Pro} & ChartQA & \shortstack{Count\\BenchQA} & CountQA & \shortstack{MME\\RealWorld\\Lite} & \shortstack{Visual\\Probe} & \textbf{Overall} & \shortstack{\textbf{Gaze}\\\textbf{Alignment}\\{[bits/fix]}} \\
    \hline
    \multicolumn{9}{@{}l}{\textit{Base model}} \\
    MiMo-VL-7B + CoT & $57.4^{\pm 0.4}$ & $89.2^{\pm 0.1}$ & $82.5^{\pm 1.6}$ & $21.9^{\pm 0.6}$ & $34.8^{\pm 0.1}$ & $33.0^{\pm 1.0}$ & $53.1^{\pm 0.3}$ & $0.04^{\pm 0.01}$ \\
    \hline
    \multicolumn{9}{@{}l}{\textit{Human-fixation SFT}} \\
    Raw-Gaze-SFT & $24.9^{\pm 0.1}$ & $66.1^{\pm 0.1}$ & $57.1^{\pm 0.4}$ & $16.2^{\pm 0.5}$ & $29.0^{\pm 0.7}$ & $35.1^{\pm 0.8}$ & $38.1^{\pm 0.2}$ & $\mathbf{0.88}^{\pm 0.04}$ \\
    Gaze-All & $53.4^{\pm 0.4}$ & $89.7^{\pm 0.1}$ & $81.5^{\pm 0.8}$ & $19.9^{\pm 0.3}$ & $49.7^{\pm 0.4}$ & $38.3^{\pm 0.6}$ & $55.4^{\pm 0.2}$ & $0.77^{\pm 0.05}$ \\
    Gaze-Subset & $\mathbf{58.8}^{\pm 0.5}$ & $\mathbf{91.0}^{\pm 0.0}$ & $88.6^{\pm 0.0}$ & $26.2^{\pm 0.4}$ & $\mathbf{52.4}^{\pm 0.5}$ & $42.8^{\pm 0.8}$ & $60.0^{\pm 0.2}$ & $0.59^{\pm 0.04}$ \\
    \modelsft{} & $57.8^{\pm 0.2}$ & $\mathbf{90.6}^{\pm 0.0}$ & $\mathbf{91.7}^{\pm 0.4}$ & $\mathbf{35.0}^{\pm 0.7}$ & $\mathbf{52.1}^{\pm 0.2}$ & $\mathbf{44.7}^{\pm 1.2}$ & $\mathbf{62.0}^{\pm 0.2}$ & $0.65^{\pm 0.05}$ \\
    \hline
    \end{tabular}%
    }
    \endgroup
    \caption{\textbf{SFT task performance and gaze alignment with MiMo-VL-7B.} Condition names follow Table~\ref{tab:gaze-supervision-strategies}. Task scores report Avg@3 ($\pm$ SEM). We bold SFT task means within 0.5 percentage points of the best SFT mean in each column, and the best SFT gaze-alignment mean.}
    \label{tab:mimo-sft-task-results}
\end{table}

\par\noindent
\begin{minipage}{\linewidth}
\paragraph{Point-budget analysis.}
Figure~\ref{fig:additional-rl-budgets} reports \modelrl{}, No-Gaze RL, and Point-Grounded RL at the point budgets used in Figure~\ref{fig:gaze-alignment-overview}b--c.

\begin{figure}[H]
    \centering
    \includegraphics[width=0.58\linewidth]{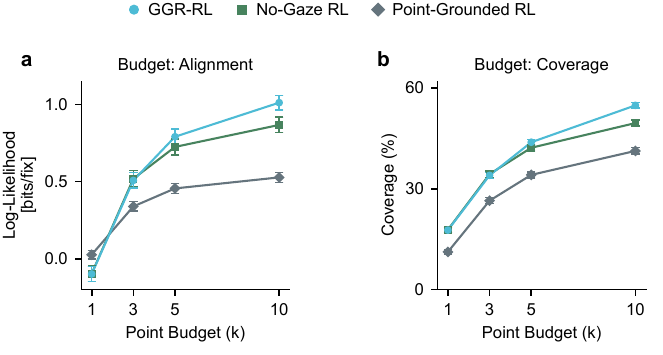}
    \caption{\textbf{Additional point-budget results.} (a) Alignment and (b) coverage for \modelrl{}, No-Gaze RL, and Point-Grounded RL using up to the first $k\in\{1,3,5,10\}$ unique output points. Metrics and axis limits match Figure~\ref{fig:gaze-alignment-overview}b--c. Reported is mean $\pm$ SEM.}
    \label{fig:additional-rl-budgets}
\end{figure}
\end{minipage}
\par

\endgroup

\begingroup
\raggedbottom
\Needspace{0.95\textheight}
\setlength{\intextsep}{3pt plus 1pt minus 1pt}
\setlength{\parskip}{4pt}
\subsection{Additional Trial Examples}

\begin{figure}[H]
    \centering
    \includegraphics[width=\linewidth,height=0.50\textheight,keepaspectratio]{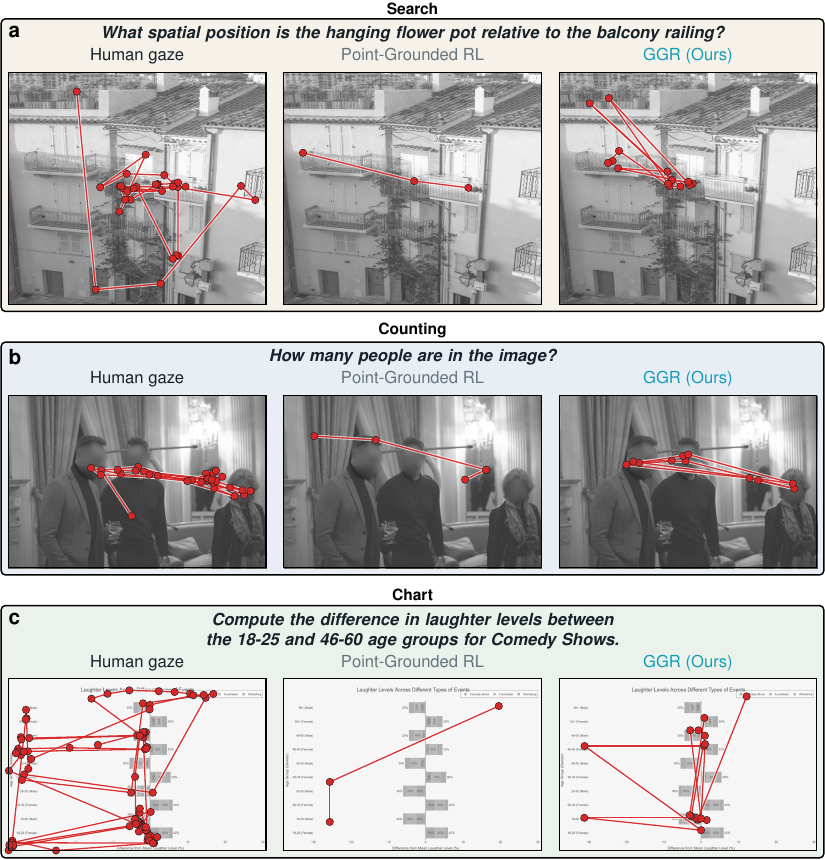}
    \caption{Human gaze and model grounding traces from Point-Grounded RL and \model{} on additional held-out examples across our three datasets.}
    \label{fig:dataset_trials}
\end{figure}

\paragraph{Examples showing internal attention tracking grounded reasoning.}
We extracted decoder-to-image attention while \modelrl{} predicted each coordinate pair, averaging across all attention heads and tokens in the pair at a fixed layer (21 for counting and 24 for the chart). We mapped these weights onto image patches, normalized each map by its maximum, and overlaid it on the original image. Black circles mark the emitted coordinates (Figure~\ref{fig:internal-attention-examples}).

\begin{figure}[H]
    \centering
    \includegraphics[width=\linewidth,height=0.24\textheight,keepaspectratio]{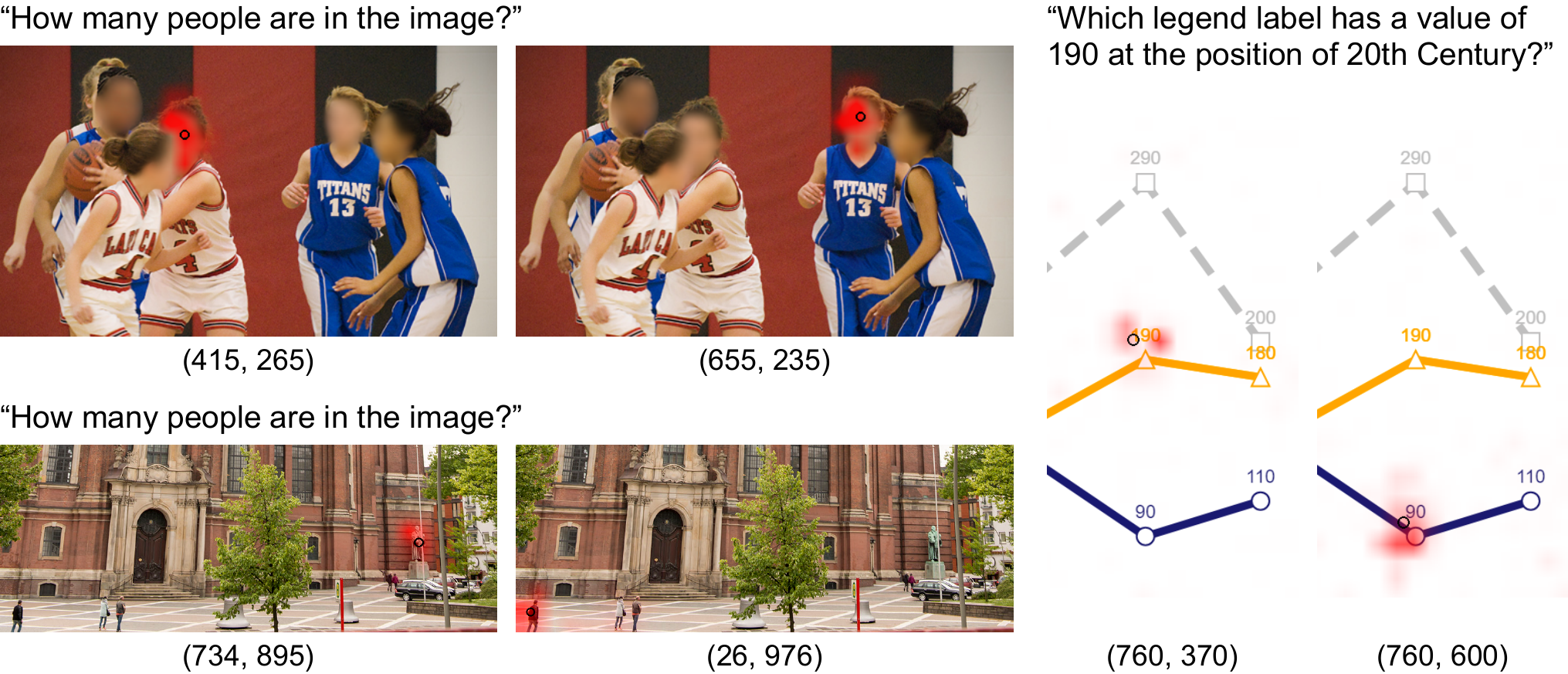}
    \caption{\textbf{Examples showing internal attention tracking grounded reasoning in \modelrl{}.} Internal attention (red) and emitted coordinates (black circles).}
    \label{fig:internal-attention-examples}
\end{figure}

\endgroup

\clearpage

\end{document}